\documentclass[10pt,journal,compsoc]{IEEEtran}

\ifCLASSOPTIONcompsoc
  \usepackage[nocompress]{cite}
\else
  \usepackage{cite}
\fi

\usepackage{graphicx}
\usepackage{tikz}
\usepackage{comment}
\usepackage{amsmath,amssymb} 
\usepackage[colorlinks,linkcolor=red]{hyperref}

\usepackage{multirow}
\usepackage{algpseudocode}  
\usepackage{amsmath,amssymb,amsthm}  
\newtheorem{corollary}{Corollary}

\usepackage{array}
\usepackage[ruled,linesnumbered]{algorithm2e}
\usepackage{hyperref}
\usepackage{colortbl}
\usepackage{amssymb}
\usepackage{bbding}

\begin{document}
%

\title{Improving Viewpoint Robustness for Visual Recognition via Adversarial Training}

%
%

\author{Shouwei~Ruan,~
        Yinpeng~Dong,~
        Hang~Su,~
        Jianteng Peng,~
        Ning Chen,~
        and Xingxing~Wei
\IEEEcompsocitemizethanks{\IEEEcompsocthanksitem Shouwei Ruan and Xingxing Wei are with the Institute of Artificial Intelligence, Beihang University, No.37, Xueyuan Road, Haidian District, Beijing, 100191, P.R. China. (E-mail: {shouweiruan, xxwei}@buaa.edu.cn).
Yinpeng Dong, Hang Su and Ning Chen are with the Institute for Artificial Intelligence, Beijing National Research Center for Information Science and Technology, Department of Computer Science and Technology, Tsinghua University, Beijing 100084, China.
Jianteng Peng is with the OPPO Intellisense and Interaction
Research Department.
\IEEEcompsocthanksitem Corresponding author: Xingxing Wei and Hang Su.}
}

%
%


\IEEEtitleabstractindextext{%
\begin{abstract}
Viewpoint invariance remains challenging for visual recognition in the 3D world, as altering the viewing directions can significantly impact predictions for the same object. While substantial efforts have been dedicated to making neural networks invariant to 2D image translations and rotations, viewpoint invariance is rarely investigated. Motivated by the success of adversarial training in enhancing model robustness, we propose Viewpoint-Invariant Adversarial Training (VIAT) to improve the viewpoint robustness of image classifiers. Regarding viewpoint transformation as an attack, we formulate VIAT as a minimax optimization problem, where the inner maximization characterizes diverse adversarial viewpoints by learning a Gaussian mixture distribution based on the proposed attack method GMVFool. The outer minimization obtains a viewpoint-invariant classifier by minimizing the expected loss over the worst-case viewpoint distributions that can share the same one for different objects within the same category. Based on GMVFool, we contribute a large-scale dataset called ImageNet-V+ to benchmark viewpoint robustness. Experimental results show that VIAT significantly improves the viewpoint robustness of various image classifiers based on the diversity of adversarial viewpoints generated by GMVFool. Furthermore, we propose ViewRS, a certified viewpoint robustness method that provides a certified radius and accuracy to demonstrate the effectiveness of VIAT from the theoretical perspective.
\end{abstract}

}

\maketitle

\IEEEdisplaynontitleabstractindextext

%
\IEEEpeerreviewmaketitle

\IEEEraisesectionheading{\section{Introduction}\label{sec:introduction}}

%
%
%
%

\IEEEPARstart {L}{earning} invariant representations is highly desirable across numerous computer vision tasks, as it contributes to the robustness of models when confronted with semantic-preserving transformations of images~\cite{bengio2013representation,goodfellow2009measuring}. Previous research efforts have focused on achieving invariance to image translation, rotation, reflection, and scaling in visual recognition models~\cite{engstrom2019exploring,zhang2019making,cohen2016group,sifre2013rotation}. However, these studies primarily address 2D image transformations and pay less attention to viewpoint transformations in the 3D domain~\cite{zemel1990discovering}. It has been observed that visual recognition models are susceptible to changes in viewpoint~\cite{alcorn2019strike,barbu2019objectnet,dong2022viewfool}, exhibiting a noticeable disparity compared to human vision, which exhibits robust object recognition across different viewpoints~\cite{biederman1987recognition}. Given the naturalness and prevalence of viewpoint variations in safety-critical areas (e.g., autonomous driving, robotics, surveillance), it becomes crucial to imbue visual recognition with viewpoint invariance.

\begin{figure}[t]
  \centering
  \includegraphics[width=0.99\linewidth]{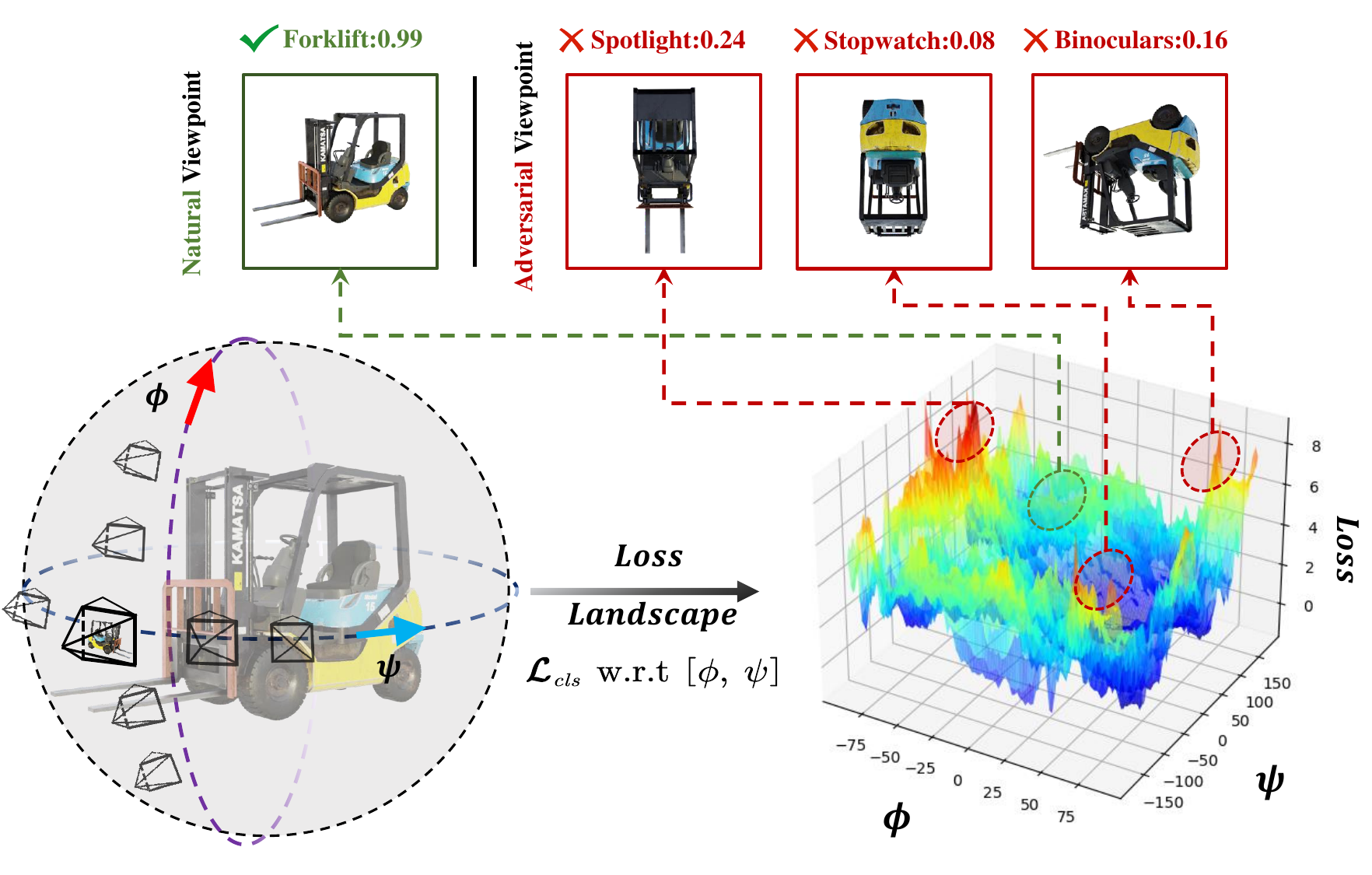}
   \caption{An illustration of viewpoint changes on model performance. We show the loss landscape w.r.t. yaw and pitch of the camera, which demonstrates multiple regions of adversarial viewpoints (We use ResNet-50 as the target model~\cite{he2016deep}). 
   }
   \vspace{-0.4cm}
   \label{fig:0}
\end{figure}

\begin{figure*}
  \centering
  \includegraphics[width=0.99\linewidth]{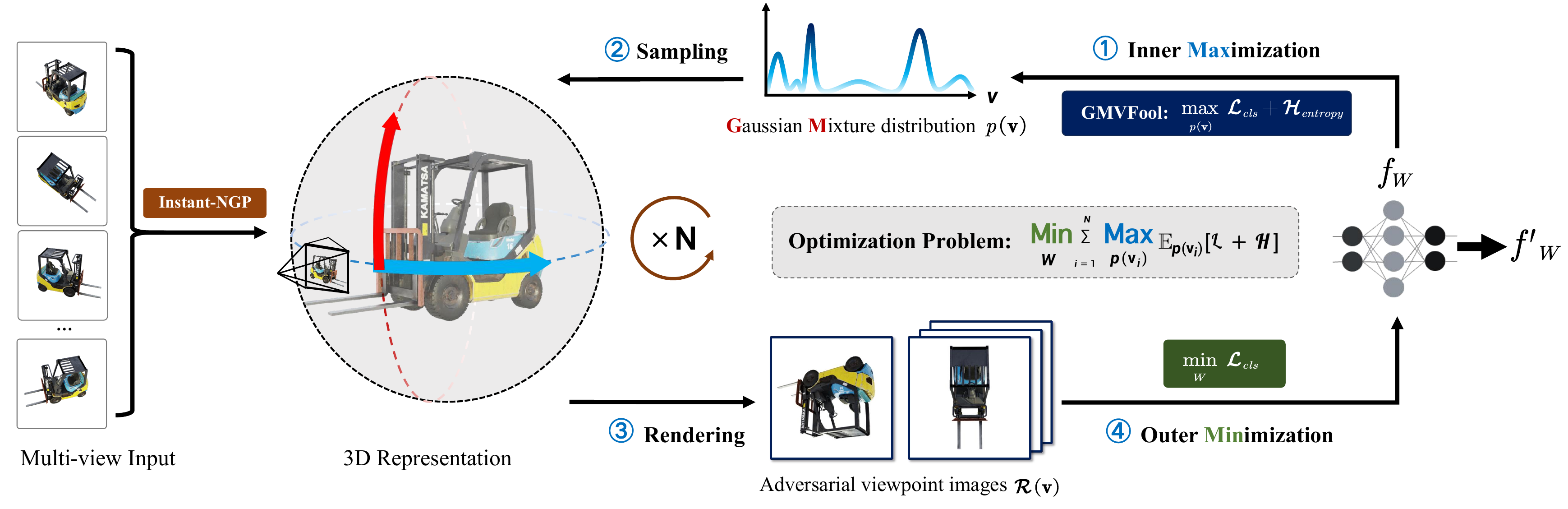}
   \caption{An overview of our VIAT framework. We first train the NeRF representation of each object given multi-view images. The inner maximization learns a Gaussian mixture distribution of adversarial viewpoints by maximizing the expectation of classification loss and entropy regularization. The outer minimization samples adversarial viewpoints from the optimized distributions and renders 2D images from adversarial viewpoints, which are fed into the network along with clean samples to train viewpoint-invariant classifiers.}
   \label{fig:framework}
   \vspace{-0.3cm}
\end{figure*}

Despite the importance, it is extremely challenging to build viewpoint-invariant visual recognition models since typical networks take 2D images as inputs without inferring the structure of 3D objects. As an effective data-driven approach, Adversarial Training (AT) augments training data with adversarially generated samples under a specific threat model and shows promise to improve model robustness against various perturbations, including additive adversarial perturbations~\cite{madry2017towards,zhang2019theoretically}, image translation and rotation~\cite{engstrom2019exploring}, geometric transformations~\cite{kanbak2018geometric}, etc. This motivates us to regard viewpoint transformation as an attack, and utilize AT to train a robust visual recognition model via the generated adversarial viewpoint examples. 

However, it is non-trivial to directly apply traditional Adversarial Training (AT) because of the following challenges. \textbf{(1)} For the inner maximization step in AT, we need to \textbf{efficiently} search for the \textbf{diverse} viewpoints for a given object, thus providing sufficient adversarial viewpoints examples for training. As a pioneering work, \emph{ViewFool}~\cite{dong2022viewfool} encodes real-world 3D objects as Neural Radiance Fields (NeRF)~\cite{mildenhall2020nerf} given multi-view images and performs black-box optimization for generating a distribution of adversarial viewpoints. Though effective, ViewFool only adopts a unimodal Gaussian distribution, which proves inadequate to characterize multiple local maxima of the loss landscape w.r.t. viewpoint changes, as depicted in Fig.~\ref{fig:0}. We verify that this can lead to overfitting of adversarial training to the specific attack. Moreover, the optimization process involved in ViewFool is time-consuming, making adversarial training intractable. Therefore, ViewFool cannot meet the challenge. \textbf{(2)} For the outer minimization step in AT, since the optimization for inner maximization may not be sufficiently efficient due to the inherent difficulty of finding diverse adversarial viewpoints, we cannot perform the inner maximization process on the whole dataset in each training epoch like the traditional AT. Therefore, a key challenge is how to design a strategy to quickly sample the available adversarial viewpoint examples from a large-scale object set in each training epoch. 
Moreover, as the training progresses, the adversarial viewpoint distribution will gradually degrade, causing the model's performance to overfit to a limited range of adversarial viewpoints catastrophically. We also need to devise further strategies to address this issue.

To solve these problems, in this paper, we propose \textbf{ Viewpoint-Invariant Adversarial Training (VIAT)}, the first framework to improve the viewpoint robustness of visual recognition models via adversarial training. As illustrated in Fig.~\ref{fig:framework}, we formulate VIAT as a distribution-based min-max problem, where the inner maximization optimizes the distribution of diverse adversarial viewpoints, while the outer minimization trains a viewpoint-invariant classifier by minimizing the expected loss over the worst-case adversarial viewpoint distributions. For the \textbf{first} challenge, to address the limitations of ViewFool, we propose \textbf{GMVFool}, a practical solution for the inner optimization process. GMVFool generates a Gaussian mixture distribution of adversarial viewpoints for each object, promoting increased diversity to alleviate adversarial training overfitting. Additionally, we employ Instant-NGP~\cite{mueller2022instant}, a fast variant of NeRF, to accelerate the objects' 3D representation. For the \textbf{second} challenge,  we don't directly adopt the traditional AT. Instead, we improve the adversarial distribution training (ADT) \cite{dong2020adversarial} framework by incorporating GMVFool's mixture Gaussian modeling, thus helping mitigate catastrophic overfitting. To further address the efficienty and overfitting problem, we introduce two simple yet effective strategies. Firstly, we adopt a \textbf{stochastic optimization strategy}, randomly selecting objects from each category to update the distribution parameters at each epoch. It improves the overall efficiency of the training process. Secondly, we design a \textbf{distribution-sharing strategy}, leveraging the observation that adversarial viewpoint distributions are transferable across objects within the same class to improve generalization. We fine-tune classifiers on a mixture of natural and adversarial viewpoint images, improving viewpoint invariance.


To train VIAT, a multi-view dataset is required. However, existing  datasets~\cite{geusebroek2005amsterdam,kanezaki2018rotationnet,reizenstein2021common,collins2022abo} often suffer from limited realism and a narrow range of viewpoints, posing less suitable for addressing this specific topic. To address this, we devote substantial efforts to creating a novel and high-quality multi-view dataset named \textbf{IM3D}, which consists of 1k typical synthetic 3D objects spanning 100 distinct classes, tailored meticulously to align with the ImageNet categories. IM3D offers several notable advantages compared to previous datasets, as detailed in Table~\ref{table: dataset comparison}: \textbf{(1)} It encompasses a wider variety of object categories, facilitating a more comprehensive evaluation of viewpoint robustness. \textbf{(2)} It exploits physics-based rendering (PBR) technology\footnote[1]{ \scriptsize A 3D modeling and rendering technique enables physically realistic effects.} to produce realistic images. \textbf{(3)} It has accurate camera pose annotations and is sampled from a spherical space, leading to better reconstruction quality and exploration of the entire 3D space. Thus, we primarily rely on IM3D for training and conducting thorough evaluations of VIAT and perform further assessments on other multi-view datasets to demonstrate the generalization of VIAT.
\begin{table}[htb]
\caption{Comparison of our multi-view dataset with others.}
\vspace{-0.2cm}
\scriptsize
\setlength\tabcolsep{5.5pt}
\renewcommand\arraystretch{1.2}
\centering
\begin{tabular}{l|c|c|c|c|c}
\hline
Dataset           & \#Objects & \#Classes & PBR  & Full 3D  & Spherical Pose \\ \hline \hline 
ALOI \cite{geusebroek2005amsterdam}              & 1K        & -        & \XSolidBrush   &\XSolidBrush & \Checkmark     \\
MIRO \cite{kanezaki2018rotationnet}         & 120       & 12        & \XSolidBrush  & \XSolidBrush & \Checkmark         \\
OOWL \cite{ho2019catastrophic}        & 500       & 25        &  \XSolidBrush   & \XSolidBrush & \XSolidBrush          \\
CO3D \cite{reizenstein2021common}              & 18.6K     & 50        &  \XSolidBrush  &  \XSolidBrush   & \XSolidBrush          \\
ABO \cite{collins2022abo}               & 8K        & 63        &  \Checkmark &  \Checkmark   & \XSolidBrush          \\
Dong~et al. \cite{dong2022viewfool}               & 100        & 85        &  \Checkmark &  \Checkmark   & \Checkmark \\
\textbf{IM3D (Ours)} & 1K        & 100       &  \Checkmark  &  \Checkmark   & \Checkmark        \\ \hline
\end{tabular}
\label{table: dataset comparison}

\end{table}

Extensive experiments are conducted to validate the effectiveness of both GMVFool and VIAT in generating adversarial viewpoints and improving the viewpoint robustness of image classifiers. The experimental results demonstrate that GMVFool characterizes more diverse adversarial viewpoints while maintaining high attack success rates, which allows VIAT significantly improves the viewpoint robustness of image classifiers, encompassing models from CNN to Transformer architectures (including large vision-language models) and shows superior performance compared with alternative baselines. To better evaluate the viewpoint robustness, we construct a new out-of-distribution (OOD) benchmark named \textbf{ImageNet-V+}, comprising nearly 100k adversarial viewpoints images generated by GMVFool, is $10\times$ larger than the previously available dataset~\cite{dong2022viewfool}. We hope to serve it as a standard benchmark for evaluating viewpoint robustness in future research endeavors.

The above experiments verify the robustness of visual recognition to viewpoint changes through empirical evaluation. However, there is a lack of certified methods for assessing the viewpoint robustness from the theoretical perspective. Therefore, we propose a viewpoint robustness certification method named \emph{ViewRS}, based on the Randomized Smoothing \cite{cohen2019certified}, to evaluate various VIAT-trained models via estimating the certified radius. The code of our work is available at \url{https://github.com/Heathcliff-saku/VIAT}.

In summary, this paper has the following contributions:

\textbf{\underline{(1)}} We propose \textbf{VIAT}, the first framework for enhancing the viewpoint robustness of visual recognition models via adversarial training. Unlike previous methods that rely on data augmentation\cite{madan2020capability} or extra regularizers\cite{yang2019invariance}, VIAT enhances viewpoint robustness by minimizing the model's loss expectation over the worst-case adversarial distribution of viewpoints. It does not require 3D object cues and only uses multi-view images as input, making it be trained easily.

\textbf{\underline{(2)}} We contribute \textbf{GMVFool}, an efficient viewpoint attack method that optimizes the Gaussian mixture distribution of adversarial viewpoints through multi-view images, which can capture a diversity of adversarial viewpoints simultaneously and achieves a 160-fold acceleration compared to the previous method ViewFool.

\textbf{\underline{(3)}} We propose \textbf{ViewRS}, a viewpoint robustness certification method based on Randomized Smoothing. ViewRS constructs a viewpoint-smoothed classifier based on the original classifier and estimates the certification radius to evaluate the provable viewpoint robustness. We employ ViewRS to demonstrate the effectiveness of VIAT and provide viewpoint robustness certifications for various models.

\textbf{\underline{(4)}} We contribute a multi-view dataset to train VIAT and a benchmark to assess the viewpoint robustness. The multi-view dataset \textbf{IM3D} contains 1k synthetic 3D objects from 100 ImageNet classes and has realistic viewpoint images. The viewpoint robustness benchmark dataset \textbf{ImageNet-V+}, containing 100K images from the adversarial viewpoints, is 10× larger than the previous ImageNet-V\cite{dong2022viewfool}. 

This paper is an extended version of our ICCV work~\cite{ruan2023towards}, where we have made significant improvements and extensions: \textbf{(1)} As mentioned above,  we propose a certified viewpoint robustness method named as \emph{ViewRS}, which extends Randomized Smoothing to the viewpoint parameters domain and constructs a viewpoint-smoothed classifier. It can support the further certification and evaluation of various models against viewpoint changes. The technical details are introduced in Sec. \ref{viewrs-s}, and the certified results w.r.t various models are given in  Sec.~\ref{sec:Certification}. \textbf{(2)} Considering the recent remarkable success of Large Vision-Language Models (VLMs) in various vision tasks, we further evaluate the performance of VLMs represented by CLIP, BLIP, etc., on viewpoint OOD  data in Sec. \ref{sec:imagenet-v+}. Despite these models being considered robust in some OOD domains, their performance remains subpar regarding adversarial viewpoints. \textbf{To the best of our knowledge, this is the first time to exploit the viewpoint robustness of LVMs}. We believe these results will provide valuable insights for advancing large models in the vision domain. \textbf{(3)} We polish the whole paper and rewrite the introduction, related works, and experiments as well as re-illustrate some figures to facilitate a better understanding of the paper. 

The rest of this paper is structured as follows: Section~\ref{sec:2} reviews the related works on adversarial samples, robustness to viewpoint transformation, and adversarial training. Section~\ref{sec:3} describes the preliminary on NeRF and the details of the proposed methods. Section~\ref{sec:4} presents and analyzes the experimental results. Section~\ref{sec:5} concludes the paper.

\section{Related Work}
\label{sec:2}
\subsection{Adversarial Samples of Deep Neural Networks}
\label{sec:asdnn}

Recent studies have revealed the vulnerability of Deep Neural Networks (DNNs) to adversarial samples~\cite{goodfellow2014explaining, szegedy2013intriguing}, which can manifest in various forms. On the one hand, earlier research has successfully deceived visual models in both digital~\cite{wei2021black,liang2021parallel,wei2022sparse} and physical domains~\cite{wei2022physically,wei2023physically} by strategically applying small perturbations within $l_p$-norms constraints to natural samples. This poses a significant threat to deploying deep networks and their applications in safety-critical domains. On the other hand, adversarial patch samples represent a more menacing type of adversarial sample, particularly concerning physical environments, leading to more effective physical manipulations. Adversarial patch samples are typically generated by introducing unconstrained perturbations to specific regions of natural samples, giving rise to various optimization paradigms, including pattern-aware optimization~\cite{karmon2018lavan, brown2017adversarial}, location-aware optimization~\cite{wei2022adversarial},  and joint optimization~\cite{wei2022simultaneously}. While these samples pose a considerable threat, they are intentionally crafted and seldom encountered in real-world scenarios. Furthermore, researchers have also investigated the susceptibility of deep networks to natural perturbations such as image rotations and translations~\cite{engstrom2019exploring}, geometric deformations\cite{kanbak2018geometric, alfarra2022deformrs}, lighting and shadowing conditions~\cite{zhong2022shadows,li2023physical}, and others. Examining these "natural adversarial samples" is crucial for narrowing the gap between human and machine vision regarding out-of-distribution generalization.

Our work specifically focuses on a fundamental and novel aspect of natural adversarial samples, namely the robustness against viewpoint transformations—whose background will be extensively explored in the following section.

\subsection{Robustness to Viewpoint Transformation}
As deep neural networks are being increasingly applied in numerous safety-critical fields, it is necessary to investigate the robustness of visual models against viewpoint transformations in 3D space. Several existing works introduce images including various uncommon camera viewpoints, object poses, and object shapes to evaluate out-of-distribution (OOD) generalization under viewpoint changes, such as the ObjectNet~\cite{barbu2019objectnet}, OOD-CV~\cite{zhao2022ood}, and ImageNet-R~\cite{hendrycks2021many} datasets. Despite significant efforts, they cannot evaluate model performance under the worst-case viewpoint transformation. Alcorn et al. first~\cite{alcorn2019strike} generate adversarial viewpoint samples for 3D objects using a differentiable renderer and find that the model is highly susceptible to viewpoint changes; Hamdi et al.~\cite{hamdi2020towards} demonstrate the effect of viewpoint perturbation on the model performance of 3D objects and use integral boundary optimization to find robust viewpoint regions for the model; Madan et al.~\cite{madan2020and} introduce diverse category-viewpoint combination images through digital objects and scenes to improve the model's generalization to OOD viewpoints. However, a common limitation of these methods is that they require the complete 3D structure of the objects. Recently, Dong et al. \cite{dong2022viewfool} further propose ViewFool, which uses NeRF to build 3D representations of objects within multi-view images and optimizes the adversarial viewpoint distribution under an entropy regularizer. But it cannot discover diverse adversarial viewpoints. 

Our work differs from them mainly in that we focus on improving the viewpoint robustness of models rather than attacking them and then design a more efficient method to generate diverse adversarial viewpoints for this purpose.

\subsection{Adversarial Training}
\label{sec:Spatial Robustness in Vision Recognition}

Introduced by Goodfellow et al.\cite{goodfellow2014explaining}, Adversarial Training (AT) is widely acknowledged as the most effective method to enhance the robustness of deep learning models~\cite{athalye2018robustness,bai2021recent}. Classical AT frameworks, such as PGD-AT~\cite{madry2017towards}, have been the basis for previous studies proposing enhancement strategies from various perspectives~\cite{tramer2017ensemble,zhang2019theoretically,pang2020bag}. Adversarial training is being found widespread use in diverse deep learning tasks, including visual recognition~\cite{goodfellow2014explaining, madry2017towards,zhang2019theoretically}, point cloud recognition~\cite{liu2019extending,wang2022art}, and text classification~\cite{miyato2016adversarial}. Regarding the application of AT on viewpoint robustness, Alcorn et al.~\cite{alcorn2019strike} demonstrate the efficacy of AT by generating adversarial viewpoint images using the renderer. However, due to relying solely on the traditional AT paradigm, this approach only improves the robustness of known objects and does not generalize to unseen objects.

The difference from our work is that we don't rely on traditional renderers and 3D information of objects and can significantly improve the model's adversarial viewpoint generalization ability for unseen objects.

\section{Methodology}
\label{sec:3}
The proposed Viewpoint-Invariant Adversarial Training (VIAT) is given here. We first introduce the background of NeRF in Sec.~\ref{sec:nerf} and the problem formulation in Sec.~\ref{sec:Formulation}, and then present the solutions of VIAT to the inner maximization in Sec.~\ref{sec:Attack} and outer minimization in Sec.~\ref{sec:Defense}. The certified method is given in Sec.~\ref{viewrs-s}. 

\subsection{Preliminary on Neural Radiance Fields (NeRF)}
\label{sec:nerf}

Given a set of multi-view images, NeRF~\cite{mildenhall2020nerf} has the ability to implicitly represent the object/scene as a continuous volumetric radiance field $F:(\mathbf{x},\mathbf{d})\rightarrow (\mathbf{c}, \tau)$, 
where $F$ maps the 3D location $ \mathbf{x}\in \mathbb{R} ^3 $ and the viewing direction $ \mathbf{d}\in \mathbb{S} ^2 $ to an emitted color $ \mathbf{c}\in [0,1]^3 $ and a volume density $\tau\in\mathbb{R}^+ $. Then, using the volume rendering with stratified sampling, we can render a 2D image from a specific viewpoint. Given a camera ray $\mathbf{r}(t) = \mathbf{o} + t\mathbf{d} $ emitted from the camera center $\mathbf{o}$ through a pixel on the image plane, the expected color $\hat{C}(\mathbf{r})$ of the pixel can be calculated by a discrete set of sampling points $\{t_m\}_{m=1}^{M}$ as
\begin{equation} \hat{C}(\mathbf{r})  = \sum_{m=1}^M T(t_m)\cdot \alpha (\tau(t_m) \cdot \delta_m) \cdot \mathbf{c}(t_m),
    \label{eq:nerf_2}
\end{equation}
where $T(t_m)  = \exp(-\sum_{j=1}^{m-1}\tau(t_j)\cdot\delta_j)$,  $\tau(t_m)$ and $\mathbf{c}(t_m)$ denote the volume density and color at point $ \mathbf{r}(t_m)$, $\delta_m = t_{m+1}-t_m$ is the distance between adjacent points, and $ \alpha{(x)}=1-\exp{(-x)} $. $F$ is approximated by a multi-layer perceptron (MLP) network and optimized by minimizing the $L_2$ loss between the rendered and ground-truth pixels.

Although NeRF can render photorealistic novel views, both training and rendering are extremely time-consuming. Instant-NGP~\cite{mueller2022instant} proposes a fast implementation of NeRF by adaptive and efficient multi-resolution hash encoding. Therefore, in this paper, we adopt Instant-NGP to accelerate the training and volumetric rendering of NeRF.

\subsection{Problem Formulation}
\label{sec:Formulation}

In visual recognition, viewpoint invariance indicates that a model $f(\cdot)$ can make an identical prediction given two views of the same object as follows:
\begin{equation}
f(I(\mathbf{v}_1))=f(I(\mathbf{v}_2)), \;\;\; \forall (\mathbf{v}_1, \mathbf{v}_2)
\end{equation}
where $I(\mathbf{v}_1)$ and $I(\mathbf{v}_2)$ are two images taken from arbitrary viewpoints $\mathbf{v}_1$ and $\mathbf{v}_2$ of the object. However, recent studies~\cite{barbu2019objectnet,dong2022viewfool,alcorn2019strike} have revealed that typical image classifiers are susceptible to viewpoint changes. As viewpoint variations in the 3D space cannot be simply simulated by 2D image transformations, it remains challenging to improve viewpoint invariance/robustness. Motivated by the success of adversarial training in improving model robustness, we propose \textbf{Viewpoint-Invariant Adversarial Training (VIAT)} by learning on worst-case adversarial viewpoints.

Formally, viewpoint changes can be described as rotation and translation of the camera in the 3D space~\cite{dong2022viewfool}. We let $\mathbf{v} = [\mathbf{R}, \mathbf{T}] \in \mathbb{R}^6$ denote the viewpoint parameters bounded in $[\mathbf{v}_{\min}, \mathbf{v}_{\max}]$, where $ \mathbf{R}=[\psi,\theta,\phi]$ is the camera rotation along the z-y-x axes using the Tait-Bryan angles, and $ \mathbf{T}=[\Delta _x,\Delta _y,\Delta_z]$ is the camera translation along the three axes.
Given a dataset $\{\mathrm{obj}_i\}_{i=1}^ N$ of $N$ objects and the corresponding ground-truth labels $\{y_i\}_{i=1}^N$ with $y_i\in\{1,...,C\}$, we suppose that a set of multi-view images is available for each object. With these images, we first train a NeRF model for each object using Instant-NGP to obtain a neural renderer that can synthesize new images from any viewpoint of the object. Rather than finding an adversarial viewpoint $\mathbf{v}_i$ for each object $\mathrm{obj}_i$, VIAT characterizes diverse adversarial viewpoints by learning the underlying distribution $p(\mathbf{v}_i)$, which can be formulated as a distribution-based minimax optimization problem:
\begin{equation}
\min_{\mathbf{W}}\!\sum_{i=1}^{N}   \max_{p(\mathbf{v}_i ) }\left [ \mathbb{E}_{p(\mathbf{v}_i )}\! \left [  \mathcal{L}\left(f_{\mathbf{W}}\left (\mathcal{R}(\mathbf{v}_i )  \right ), y_i\right )\right ]\! + \!\lambda\! \cdot\!\mathcal{H}(p(\mathbf{v}_i))\right ]\!,
\label{eq: AT formulation}
\end{equation}
where $\mathbf{W}$ denotes the parameters of the classifier $f_{\mathbf{W}}$, $\mathcal{R}(\mathbf{v}_i)$ is the rendered image of the $i$-th object given the viewpoint $\mathbf{v}_i$, $\mathcal{L}$ is a classification loss (e.g., cross-entropy loss), and $\mathcal{H}(p(\mathbf{v}_i))=-\mathbb{E}_{p(\mathbf{v}_i)}[\log p(\mathbf{v}_i)]$ is the entropy of the distribution $p(\mathbf{v}_i)$ to avoid the degeneration problem and help to capture more diverse adversarial viewpoints~\cite{dong2022viewfool}.

As can be seen in Eq.~\eqref{eq: AT formulation}, the inner maximization aims to learn a distribution of adversarial viewpoints under an entropic regularizer, while the outer minimization aims to optimize model parameters by minimizing the expected loss over the worst-case adversarial viewpoint distributions. The motivation of using a distribution instead of a single adversarial viewpoint for adversarial training is two-fold. First, learning a distribution of adversarial viewpoints can effectively mitigate the reality gap between the real objects and their neural representations~\cite{dong2022viewfool}. Second,
the distribution is able to cover a variety of adversarial viewpoints to alleviate potential overfitting of adversarial training, leading to better generalization performance. 

To solve the minimax problem, a general algorithm is to first solve the inner problem and then perform gradient descent for the outer problem at the inner solution in a sequential manner based on the Danskin's theorem~\cite{danskin2012theory}. Next, we introduce the detailed solutions to the inner maximization and outer minimization problems, respectively.

\subsection{Inner Maximization: GMVFool}
\label{sec:Attack}

The key to the success of VIAT in Eq.~\eqref{eq: AT formulation} is the solution to the inner maximization problem. A natural way to solve the problem is to parameterize the distribution of adversarial viewpoints with trainable parameters. The previous method ViewFool~\cite{dong2022viewfool} adopts a unimodal Gaussian distribution and performs black-box optimization based on natural evolution strategies (NES)~\cite{wierstra2014natural}. However, due to the insufficient expressiveness of the Gaussian distribution, ViewFool is unable to characterize multiple local maxima of the loss landscape w.r.t. viewpoint changes, as shown in Fig.~\ref{fig:0}. Thus, performing adversarial training with ViewFool is prone to overfitting to the specific attack and leads to poor generalization performance, as validated in the experiment. To alleviate this problem, we propose \textbf{GMVFool}, which learns a Gaussian mixture distribution of adversarial viewpoints to cover multiple local maxima of the loss landscape for more generalizable adversarial training. 

For the sake of simplicity, we omit the subscript $i$ in this subsection since the attack algorithm is the same for all objects. Specifically, we parameterize the distribution $p(\mathbf{v})$ by a mixture of $K$ Gaussian components and take the transformation of random variable approach to ensure that the support of $p(\mathbf{v})$ is contained in $[\mathbf{v}_{\min},\mathbf{v}_{\max}]$ as:
\begin{equation}\label{eq:dis}
 \mathbf{v} = \mathbf{a}\cdot \tanh(\mathbf{u})+\mathbf{b}, \; p(\mathbf{u}|\Psi) = \sum_{k=1}^{K} \mathbf{\omega}_k\mathcal{N} (\mathbf{u}|\boldsymbol{\mu}_k,\boldsymbol{\sigma}^2_k\mathbf{I}),   
\end{equation}
where $\mathbf{a}  = (\mathbf{v}_{\max}-\mathbf{v}_{\min})/2$, $\mathbf{b}  = (\mathbf{v}_{\max}+\mathbf{v}_{\min})/2$, $\Psi=\{\omega_k, \boldsymbol{\mu}_k, \boldsymbol{\sigma}_k\}^K_{k=1}$ are the parameters of the mixture Gaussian distribution with weight $\omega_k\in [0,1]$ ($\sum_{k=1}^K\mathbf{\omega }_k=1$), mean $\boldsymbol{\mu}_k\in \mathbb{R} ^{6}$ and standard deviation $\boldsymbol{\sigma }_k\in \mathbb{R} ^{6}$ of the $k$-th Gaussian component. Note that in Eq.~\eqref{eq:dis}, $\mathbf{u}$ actually follows a mixture Gaussian distribution while $\mathbf{v}$ is obtained by a transformation of $\mathbf{u}$ for proper normalization.  

Now the probability density function $p(\mathbf{u}|\Psi)$ is in the summation form, which is hard to calculate the gradients. Thus, we introduce a latent one-hot vector $\boldsymbol{\Gamma}=[\gamma_1, ..., \gamma_K]$ determining which Gaussian component the sampled viewpoint belongs to, and obeying a multinomial distribution with probability $\omega_k$, as $p(\boldsymbol{\Gamma}|\Psi)=\prod_{k=1}^{K} \mathbf{\omega}_k^{\gamma_k}$. 
With the latent variables, we represent  $p(\mathbf{u}|\Psi)$ as a multiplication form with $\boldsymbol{\Gamma}$ as $p(\mathbf{u},\boldsymbol{\Gamma}|\Psi) = \prod_{k=1}^{K} \mathbf{\omega}_k^{\gamma_k}\mathcal{N}(\mathbf{u}|\boldsymbol{\mu}_k,\boldsymbol{\sigma}^2_k\mathbf{I})^{\gamma_k}$
and $p(\mathbf{u}|\Psi)=\sum_{\Gamma}p(\mathbf{u},\boldsymbol{\Gamma}|\Psi)$, which is convenient for taking derivatives w.r.t. distribution parameters $\Psi$. 

Given the parameterized distribution $p(\mathbf{v})$ defined in Eq.~\eqref{eq:dis}, the inner maximization problem of  Eq.~\eqref{eq: AT formulation} becomes:
\begin{equation}
    \begin{split}
        \max_{\Psi}\; \mathbb{E}_{ p(\mathbf{u},\boldsymbol{\Gamma}|\Psi)} & \big[ \mathcal{L}(f_\mathbf{W}  (\mathcal{R}(\mathbf{a}\cdot \tanh(\mathbf{u})+\mathbf{b}) ) ,y) \\
        & - \lambda\cdot \log p(\mathbf{a}\cdot \tanh(\mathbf{u})+\mathbf{b})\big],
    \end{split}
    \label{eq: attack2}
\end{equation}
where the second term is the negative log density, whose expectation is the distribution's entropy (see \textbf{Appendix A}). 

To solve this optimization problem, we adopt gradient-based methods to optimize the distribution parameters $\Psi$.
To back-propagate the gradients from random samples to the distribution parameters, we can adopt the low-variance reparameterization trick~\cite{kingma2013auto}. Specifically, we reparameterize $\mathbf{u}$ as 
$\mathbf{u} =  {\prod_{k=1}^{K}}\boldsymbol{\mu}^{\gamma_k}_k+{ \prod_{k=1}^{K}}\boldsymbol{\sigma}^{\gamma_k}_k\cdot\mathbf{r}$, where $\mathbf{r}\sim \mathcal{N}(\mathbf{0},\mathbf{I})$. 
With this reparameterization, the gradients of the loss function in Eq.~\eqref{eq: attack2} w.r.t. $\Psi$ can be calculated. However, similar to ViewFool, although the rendering process of NeRF is differentiable, it requires significant memory overhead to render the full image. Thus, we also resort to NES to obtain the natural gradients of the classification loss with only query access to the model. For the entropic regularizer, we directly compute its true gradient. Therefore, the gradients of the objective function in Eq.~\eqref{eq: attack2} w.r.t. $\omega_k, \boldsymbol{\mu}_k$ and $\boldsymbol{\sigma}_k$ can be derived as (proof in \textbf{Appendix A}):
\begin{equation}\small
    \begin{split}
    \nabla_{\omega_k} = \mathbb{E}_{\mathcal{N}(\mathbf{r}|\mathbf{0},\mathbf{I})} & \left \{   \gamma_k \cdot \left [\mathcal{L}_\text{cls} \cdot \frac{1}{\omega _k} - \lambda \right ]\right \}; \\
    \nabla_{\boldsymbol{\mu}_k} = \mathbb{E}_{\mathcal{N}(\mathbf{r}|\mathbf{0},\mathbf{I})}& \left \{\gamma _k \cdot \left [\mathcal{L}_\text{cls} \cdot \frac{\boldsymbol{\sigma}_k \mathbf{r} }{\omega _k} - \lambda \!\cdot\! 2 \tanh(\boldsymbol{\mu} _k+\boldsymbol{\sigma}_k \mathbf{r}  )\right]\!\right \}\!; \\
    \nabla_{\boldsymbol{\sigma}_k} = \mathbb{E}_{\mathcal{N}(\mathbf{r}|\mathbf{0},\mathbf{I})}&\left \{\gamma _k \cdot \left [\mathcal{L}_\text{cls} \cdot \frac{\boldsymbol{\sigma_k} (\mathbf{r}^2-1) }{2\omega_k} \right.\right.\\
    &\left.\left. + \lambda \cdot \frac{(1-2\mathbf{r}\cdot \tanh(\boldsymbol{\mu} _k+\boldsymbol{\sigma}_k \mathbf{r}  )\cdot{\boldsymbol{\sigma}_k}  }{\boldsymbol{\sigma}_k} \right] \right \}; \\
    \mathcal{L}_\text{cls} = \mathcal{L}(f_\mathbf{W}(\mathcal{R}(&\mathbf{a} \cdot \tanh({\prod_{k=1}^{K}}\boldsymbol{\mu}^{\gamma_k}_k+{ \prod_{k=1}^{K}}\boldsymbol{\sigma}^{\gamma_k}_k\cdot\mathbf{r})+\mathbf{b})),y).
    \end{split}
    \label{eq: 12}
\end{equation}
In practice, we use the Monte Carlo method to approximate the expectation in gradient calculation and use iterative gradient ascent to optimize the distribution parameters of each Gaussian component. In addition, we normalize $\omega_k$ after each iteration to satisfy $ {\textstyle \sum_{k=1}^{K}} \omega_k=1$. Algorithm~\ref{alg1} outlines the overall algorithm of GMVFool.

\begin{algorithm}[htb]
\small
 \caption{\small GMVFool Algorithm}	\label{alg1}
 \KwIn{ Image classifier $f_\mathbf{W}$, rendering function $\mathcal{R}$, true label $y$, number of iterations $T$, number of Monte Carlo samples $q$, learning rate $\eta$,  number of Gaussian components $K$, and balance hyperparameter $\lambda$.}
 
Initialize the Gaussian mixture distribution parameters of the object $\Psi^0=\{\omega^0_k , \boldsymbol{\mu}^0_k, \boldsymbol{\sigma}^0_k \}_{k=1}^K$\;
\For{$t=1$ to $T$}{
Sample $\{\mathbf{r}_j\}_{j=1}^q$ from $\mathcal{N}(\mathbf{0},\mathbf{I})$\;
Sample $\{\boldsymbol{\Gamma}_j\}_{j=1}^q$ from the multinomial distribution with probability $\omega^t_k$\;
Calculate $\{\mathbf{u}_j\}_{j=1}^q$\;
Calculate $\nabla_{\Psi^t}=\{\nabla_{\omega^t_k},\nabla_{\boldsymbol{\mu}^t_k},\nabla_{\boldsymbol{\sigma}^t_k}\}$ by Eq.~\eqref{eq: 12}\;
Update the parameters: $\Psi^{t+1} \leftarrow \Psi^{t}+\eta \cdot \nabla_{\Psi^t}$\;
Normalize $\omega^{t+1}_k \leftarrow \omega^{t+1}_k/ {\textstyle \sum_{k=1}^{K}}\omega^{t+1}_k$\;
}

\KwOut {Parameters of adversarial viewpoint distribution: $\Psi^T=\{\omega^T_k , \boldsymbol{\mu}^T_k, \boldsymbol{\sigma}^T_k\}_{k=1}^K$. } 
\end{algorithm}

\subsection{Outer Minimization}
\label{sec:Defense}

In the outer minimization step of Eq.~\eqref{eq: AT formulation}, our objective is to minimize the loss expectation over the learned distributions of adversarial viewpoints. However, traditional adversarial training encounters two main challenges: inefficiency and overfitting. To tackle these issues, we propose the integration of two strategies: the \textbf{stochastic update strategy} and the \textbf{distribution sharing strategy}.

\textbf{Stochastic update strategy.}
Although we introduce the efficient Instant-NGP for 3D representation, the inner maximization process still requires multiple gradient steps to converge when rendering images from new viewpoints. This poses a significant challenge for adversarial training since each optimization step in the outer minimization phase needs to solve the inner maximization problem for a batch of objects. In order to accelerate adversarial training, we present the stochastic update strategy for the inner problem. Initially, we perform inner optimization to generate adversarial viewpoint distributions for all objects $\{\text{obj}_i\}_{i=1}^N=[\{\text{obj}_{j,c}\}_{j=1}^{n_c}]_{c=1}^C$ within an acceptable number of iterations $T_0$ utilizing a pre-trained image classifier:
\begin{equation}
\text{Dist}_\text{init}=[~\{\Psi_{j,1}^{T_0} \}_{j=1}^{n_1},...,\{\Psi_{j,c}^{T_0} \}_{j=1}^{n_c},...,\{\Psi_{j,C}^{T_0} \}_{j=1}^{n_C}~].
\label{eq: dist pool}
\end{equation}
This process will take a considerable amount of time to optimize the "coarse" initial adversarial viewpoint distributions $\text{Dist}_\text{init}$ for each object in the full dataset. Next, for each fine-tuning epoch $q$, we update the distribution parameters for only part of the objects with randomly selected index $\mathcal{O}^{q}$ for each category while keeping the parameters for other objects unchanged:
\begin{equation}
\text{Dist}^{q}=\{\Psi_{j\in\mathcal{O}^{q},c}^{T_{q-1}+T^\prime}\}_{c=1}^C ~\cup~ \{\Psi_{j\notin\mathcal{O}^{q},c}^{T_{q-1}}\}_{c=1}^C,
\label{eq: dist pool2}
\end{equation}
where $T_{q-1}$ denoted the total inner iterations before epoch $q$, $T^\prime$ is the inner  iteration steps in each finetune epoch. Noted that all objects can be effectively optimized within multiple epochs. 
The rationale behind this approach is that GMVFool can learn a sufficiently wide range of adversarial viewpoints, ensuring the effectiveness of the distribution over an extended period. This strategy greatly enhances training efficiency and enables feasible adversarial training.

\begin{figure}[t]
  \centering
  \includegraphics[width=0.93\linewidth]{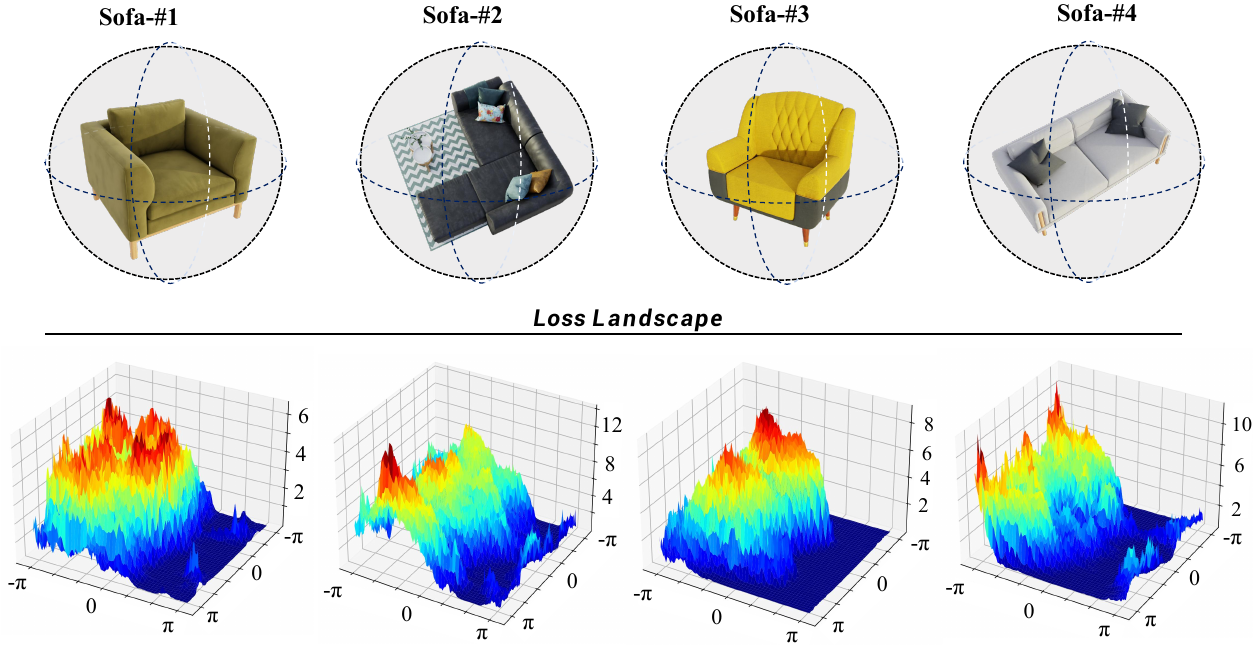}
   \caption{The adversarial viewpoint regions of objects within the same class are similar. We show the \textbf{loss landscape} w.r.t. $\psi$ and $\phi$ of different sofas based on ResNet-50, and keep $[\theta, \Delta_x, \Delta_y, \Delta_z]$ as $[0, 0, 0, 0]$.}
   \vspace{-0.3cm}
   \label{fig: transferable}
\end{figure}

\textbf{Distribution sharing strategy.} We observe that as the number of training epochs increases, the learned adversarial viewpoint distributions tend to degenerate. In other words, during the late stages of training, the diversity of adversarial viewpoints diminishes, resulting in overfitting during adversarial training and producing subpar results. To mitigate this issue, we propose a distribution sharing strategy, which involves sharing the distribution parameters among different objects within the same category. This strategy is based on our finding that the adversarial viewpoint distributions of objects within the same class exhibit a high degree of similarity, as depicted in Fig.~\ref{fig: transferable}. For each object during the sampling process, we select its own distribution parameters or randomly choose parameters from another object's distribution with the same category based on a probability $\pi$. Specifically, we first sample a random variable $\delta$ from a uniform distribution $\eth  \sim \mathcal{U}[0,1]$, and select the distribution parameters $\hat{\Psi}_{j,c}$ for sampling the viewpoints from $\text{obj}_{i,c}$ in the outer process based on the following rules:
\begin{equation}
\hat{\Psi}_{j,c} =
\begin{cases}
\Psi_{j,c}, & \eth  \le \pi,\\
\Psi_{\forall l\ne j,c}, & \eth  > \pi.
\end{cases}
\label{eq: dist pool3}
\end{equation}
A similar technique is applied in  adversarial rotation training for point clouds~\cite{wang2022art}, with the difference being that \cite{wang2022art} directly shares adversarial parameters across different categories, whereas we adopt distribution sharing within the same category for adversarial viewpoint distributions. In comparison, our distribution sharing strategy takes into account the \textbf{category-dependency} of adversarial viewpoints, and the sharing based on distributions helps mitigate the adverse effects caused by direct parameter transfer.

\textbf{Training process of VIAT.} It can be summarized as follows: during each fine-tuning epoch, the parameters of the adversarial viewpoint distribution are updated using GMVFool. Initially, all objects' parameters are optimized in the first epoch, and in each subsequent epoch, the parameters for objects within each class are randomly updated. Next, the adversarial viewpoints are sampled from a distribution based on the sharing probability. Subsequently, the corresponding adversarial examples are generated using Instant-NGP. These adversarial examples, along with clean natural samples from ImageNet, are combined to form the current training minibatch $\mathcal{B}^\prime$, which is then fed into the classifier to calculate the cross-entropy loss $\mathcal{L}_{\text{CE}}$. We utilize stochastic gradient ascent to update the weights $\mathbf{W}$ of the classifier as follows:
\begin{equation}
\mathbf{W}^{q} = \mathbf{W}^{q-1}-\eta_{\mathbf{W}} \cdot \Sigma_{(\mathbf{x}_i, y_i)\in \mathcal{B}^\prime}[\nabla_{\mathbf{W}^{q-1}} \mathcal{L}_{\text{CE}}(f_{\mathbf{W}^{q-1}}(\mathbf{x}_i),y_i)],
\label{eq:outer update}
\end{equation}
where $\eta_{\mathbf{W}}$ is the learning rate. Finally, the network parameters are optimized to obtain a viewpoint-invariant model. Algorithm~\ref{alg2} outlines the training process of VIAT. VIAT boasts acceptable time costs, which can be attributed to three factors: \textbf{\underline{(1)}} It optimizes the mixture distribution of adversarial viewpoints instead of individual ones, allowing for the generation of diverse adversarial viewpoints through distribution sampling instead of multiple optimizations. \textbf{\underline{(2)}} Using the efficient Instant-NGP accelerates the optimization of adversarial viewpoint distribution and rendering of viewpoint samples. \textbf{\underline{(3)}} The stochastic optimization strategy based on distribution transferability further reduces time consumption in adversarial training.

\begin{algorithm}[t]
\small
 \caption{\small VIAT Algorithm}	\label{alg2}
 \KwIn{ object dataset $\mathcal{D}$, clean image dataset $\mathcal{D}_{\text{clean}}$,  Image classifier $f_\mathbf{W}$, rendering function $\mathcal{R}$, training epochs $Q$, initial inner iterations $T_0$,  inner iterations in subsequent epochs $T^\prime$, number of Monte Carlo samples $q$, learning rate $\eta_{\mathbf{W}}$,  number of Gaussian components $K$, balance hyperparameter $\lambda$ and sharing probability $\pi$.}
 \tcc {distribution initialize}

Obtain $\text{Dist}_{\text{init}}$ for $\mathcal{D}$ by GMVFool with iteration $T_0$ \;

\For{$q=1$ to $Q$}{
\tcc{inner maximization}
\For{$\mathrm{each}~ \mathrm{minibatch}~\mathcal{B}\subset  \mathcal{D}$}{
\For{$\mathrm{each}~ \mathrm{input}~(\mathrm{obj}_i, y_i)\in\mathcal{B}$} {
Using GMVFool with iteration $T^\prime$ to stochastic update $\text{Dist}^q$ following Eq.~\eqref{eq: dist pool2} \;
}
}
\tcc{rendering adversarial examples}
Training minibatch $\mathcal{B^\prime} \leftarrow \mathcal{B}_{\text{clean}} \subset \mathcal{D}_{\text{clean}}$ \;
\For{$\mathrm{each}~ \mathrm{input}~(\mathrm{obj}_i, y_i)\in\mathcal{B}$} {
Sample $\eth  \sim \mathcal{U}[0,1]$\;
Choose sharing distribution $\hat{\Psi}_{i}$ for $\mathrm{obj}_i$ following Eq.~\eqref{eq: dist pool3} \;
$\mathcal{B^\prime} \leftarrow \mathcal{B^\prime} \cup  \mathcal{R}(\mathbf{v}_i)~\text{where}~\mathbf{v}_i$  is sampled from adversarial distribution $\hat{\Psi}_{i}$ following Eq.~\eqref{eq:dis}
}
\tcc{outer minimization}
Update $\mathbf{W}$ with stochastic gradient descent following Eq.~\eqref{eq:outer update}\;
}
\KwOut {Viewpoint-invariant classifier  $f_{\mathbf{W}^Q}$. } 
\end{algorithm}

\subsection{Certified Viewpoint Robustness: ViewRS}
\label{viewrs-s}
In this section, we propose ViewRS, a probabilistic certification method specialized for visual recognition models against 3D viewpoint transformations. ViewRS is based on Randomized Smoothing (RS)~\cite{cohen2019certified}, an effective and scalable approach for probabilistic certification. Therefore, we first provide an introduction to the background of RS:

\textbf{Preliminaries on Randomized Smoothing.} Randomized Smoothing is developed as a robustness certification method for DNN against the $l_p$-norm perturbations. For any classifier $f(\cdot):\mathbb{R}^d\rightarrow \mathcal{P}(\mathcal{Y})$ that maps inputs $\mathbf{x}\in \mathbb{R}^d$ to the probability simplex $\mathcal{P}(\mathcal{Y})$ over the
set of labels $\mathcal{Y}$, RS constructs a smoothed classifier $g(\cdot):\mathbb{R}^d\rightarrow \mathcal{P}(\mathcal{Y})$, which outputs the expected classification when $f$'s input undergoes additive perturbations $\boldsymbol{\epsilon}$ that sampled from a particular distribution $\mathcal{K}$:
\begin{equation}
    g(\mathbf{x}) = \mathbb{E}_{\epsilon \sim \mathcal{K}}[f(\mathbf{x}+\boldsymbol{\epsilon})].
\end{equation}
Previous studies have demonstrated that when $\mathcal{K}$  follows a Gaussian distribution~\cite{zhai2020macer} or uniform distribution~\cite{yang2020randomized}, $g(x) $ can exhibit a fixed prediction under certain conditions on the perturbation magnitude. 

While RS can be effectively performed to certify $l_p$-norm additive perturbations on images, general RS methods cannot handle specific non-additive perturbations (e.g., shadowing, geometric deformations, viewpoint changes, etc.). To address this, a significant work is DeformRS~\cite{alfarra2022deformrs}, which extends RS to certify against various 2D deformations of images by introducing a parametric-domain smooth classifier. Similarly, 3DeformRS~\cite{perez20223deformrs} adopts the design of DeformRS-PAR to certify point cloud recognition networks against point cloud deformations. As viewpoint changes can also be parameterized (as described in Sec~\ref{sec:Formulation}), the proposed ViewRS follows the relevant implications of DeformRS-PAR, providing a certification method for the robustness of classifiers against image viewpoint transformations. Precisely, given a parametric deformation function $s_\hbar$ with parameter $\hbar$, for a classifier $f(\cdot):\mathbb{R}^d\rightarrow \mathcal{P}(\mathcal{Y})$, a general parametric-domain smooth classifier can be defined as:
\begin{equation}
    g_\hbar(\mathbf{x}, \mathbf{p}) = \mathbb{E}_{\epsilon \sim \mathcal{K}}[f(I_T(\mathbf{x}, \mathbf{p}+s_{\hbar+\epsilon} )],
\end{equation}
where $I_T$ is the interpolation function and $\mathbf{p}$ denote the coordinates of inputs. Then, we have the following corollary:

\begin{corollary}  \label{corollary:1}
(restated from~\cite{alfarra2022deformrs}). Suppose $g_\hbar$ assigns the class $c_A$ for an input $\mathbf{x}$, i.e., $c_A = \arg \max_i g_\hbar^i(\mathbf{x}, \mathbf{p})$ with:
\begin{equation}
    p_A = g_\hbar^{c_A}(\mathbf{x}, \mathbf{p})~~\text{and}~~p_B = \max_{c\ne c_A}g_\hbar^{c_A}(\mathbf{x}, \mathbf{p}).
\end{equation}
Then $\arg \max_c g_{\hbar+\xi}^c(\mathbf{x}, \mathbf{p})=c_A$ for all parametric domain
 perturbations $\xi$ satisfying:

 \begin{equation}
 \begin{split}
&\left \| \xi \right \| _1 \le \beta (p_A-p_B) \quad \quad \quad \quad \quad \quad  ~ ~~\text{for}~\mathcal{K} = \mathcal{U}[-\beta, \beta], \\
&\left \| \xi \right \|_2 \le \frac{\tilde{\sigma}}{2} (\Upsilon^{-1}(p_A)-\Upsilon^{-1}(p_B) )\quad \text{for}~\mathcal{K}=\mathcal{N}(0, \tilde{\sigma}^2\mathbf{I}).
\end{split}
 \end{equation}
\end{corollary}
\noindent where $\Upsilon^{-1}(\cdot)$ is the inverse CDF of the standard Gaussian. In summary, Corollary 1 specializes DeformRS in the case of parametric perturbations. It provides an important conclusion: as long as the $l_1$ and $l_2$ norm of perturbations on parameters is sufficiently small (within the certification radius limit), $g_\hbar(\cdot)$ can maintain a constant prediction. 
\begin{table*}[t]
\caption{The \textbf{classification accuracy} from evaluation protocols with ResNet-50 and ViT-B/16, which are trained via ImageNet subset only (standard-trained), data augmentation by natural and random viewpoint images, and VIAT framework with ViewFool and GMVFool.}
\vspace{-0.3cm}
\setlength\tabcolsep{9.3pt}
\renewcommand\arraystretch{1.2}
\centering
\begin{tabular}{c|c|ccc|ccc}
\hline
\multirow{2}{*}{}        & \multirow{2}{*}{Method} & \multicolumn{3}{c|}{ResNet-50}                                                                                      & \multicolumn{3}{c}{ViT-B/16}                                                                                           \\ \cline{3-8} 
                         &                         & \multicolumn{1}{c|}{ImageNet} & \multicolumn{1}{c|}{ViewFool}       & GMVFool        & \multicolumn{1}{c|}{ImageNet} & \multicolumn{1}{c|}{ViewFool}       & GMVFool        \\ \hline \hline
Standard-trained                 & -                       & \multicolumn{1}{c|}{85.60\%}    & \multicolumn{1}{c|}{8.28\%}           & 8.98\%           & \multicolumn{1}{c|}{92.88\%}    & \multicolumn{1}{c|}{25.70\%}          & 29.10\%          \\ \hline
\multirow{2}{*}{Augmentation} & Natural                 & \multicolumn{1}{c|}{85.76\%}   & \multicolumn{1}{c|}{16.52\%}          & 19.30\%          & \multicolumn{1}{c|}{92.78\%}   & \multicolumn{1}{c|}{43.32\%}          & 46.48\%          \\
                         & Random                  & \multicolumn{1}{c|}{85.82\%}    & \multicolumn{1}{c|}{34.80\%}          & 33.52\%          & \multicolumn{1}{c|}{92.78\%}    & \multicolumn{1}{c|}{62.03\%}          & 67.34\%          \\ \hline
\multirow{2}{*}{VIAT (Ours)}    & ViewFool                & \multicolumn{1}{c|}{85.66\%}    & \multicolumn{1}{c|}{55.12\%}          & 58.75\%          & \multicolumn{1}{c|}{92.70\%}   & \multicolumn{1}{c|}{79.53\%}          & 82.03\%          \\
                         & GMVFool                 & \multicolumn{1}{c|}{85.70\%}   & \multicolumn{1}{c|}{\textbf{59.84\%}} & \textbf{59.61\%} & \multicolumn{1}{c|}{92.56\%}  & \multicolumn{1}{c|}{\textbf{82.81\%}} & \textbf{83.13\%} \\ \hline
\end{tabular}
\vspace{-0.3cm}
\label{table:defense}
\end{table*}
\textbf{Extension to viewpoint transformation.} We specialize the results of Corollary 1 to viwepoint transformations.  Note that the interpolation function $I_T$, while essential in general image deformations, is not sufficient for 3D transformations because we cannot obtain the explicit mapping relationship of  3D viewpoint changes from a image input. Since we use NeRF's implicit representation to capture the pixel values of an object $\text{obj}_i$ from parameterized viewpoint $\mathbf{v}_i \in \mathbb{R}^6$, we replace $I_T$  directly with the rendering function  $\mathcal{R}:\mathbb{R}^6 \rightarrow [0,1]^n$ proposed in Sec~\ref{sec:Formulation}. The construction of the smoothed classifier in this scenario is as follows:
\begin{equation}
    \hat{g}_\hbar(\mathbf{x},\mathbf{v}_i) = \mathbb{E}_{\epsilon \sim \mathcal{K}}[f(\mathcal{R}(\textbf{v}_i+s_{\hbar+\epsilon})].
\label{eq:viewrs}
\end{equation}
Since in this case where $s_\hbar=\hbar \in \mathbb{R}^6$, we setting $\mathbf{v}_i^\prime=\mathbf{v_i}+\hbar$. We further rewrite Eq.~\ref{eq:viewrs} as:
\begin{equation}
    \hat{g}_\hbar(\mathbf{x},\mathbf{v}_i) = \mathbb{E}_{\epsilon \sim \mathcal{K}}[f(\mathcal{R}(\textbf{v}_i^\prime+\epsilon)].
\label{eq:viewrs2}
\end{equation}
The smoothed classifier of ViewRS $\hat{g}_\hbar(\cdot)$ inherits the structure of the parameter domain smoothed classifier $g_\hbar(\cdot)$ and is certifiable against the perturbations in viewpoint parameters through Corollary~\ref{corollary:1}. In practice, we construct the smoothed classifier based on the NeRF-based 3D representation and the parameterization of viewpoint transformations proposed earlier. We follow the standard practices and utilize the public implementation from~\cite{cohen2019certified} for estimating the \textbf{Average Certification Radius (ACR)}: $\tilde{\sigma}\Upsilon^{-1}(p_A)$ using Monte Carlo sampling. As we employ Gaussian smoothing, the certification we obtained is in the $l_2$ sense. The ACR metric provides certified robustness bounds for models, indicating the range of perturbations within which the smoothed classifier  $\hat{g}_\hbar(\cdot)$, constructed from the classifier $f(\cdot)$, maintains its certified robustness. Thus, we aim to utilize the obtained ACR to further evidence the effectiveness of VIAT in achieving viewpoint invariance and provide certification of viewpoint robustness for various models.

\section{Experiments}
\label{sec:4}

\begin{figure}[b]
\vspace{-0.3cm}
	\centering
	\begin{minipage}{0.47\linewidth}
		\centering
		\includegraphics[width=0.99\linewidth]{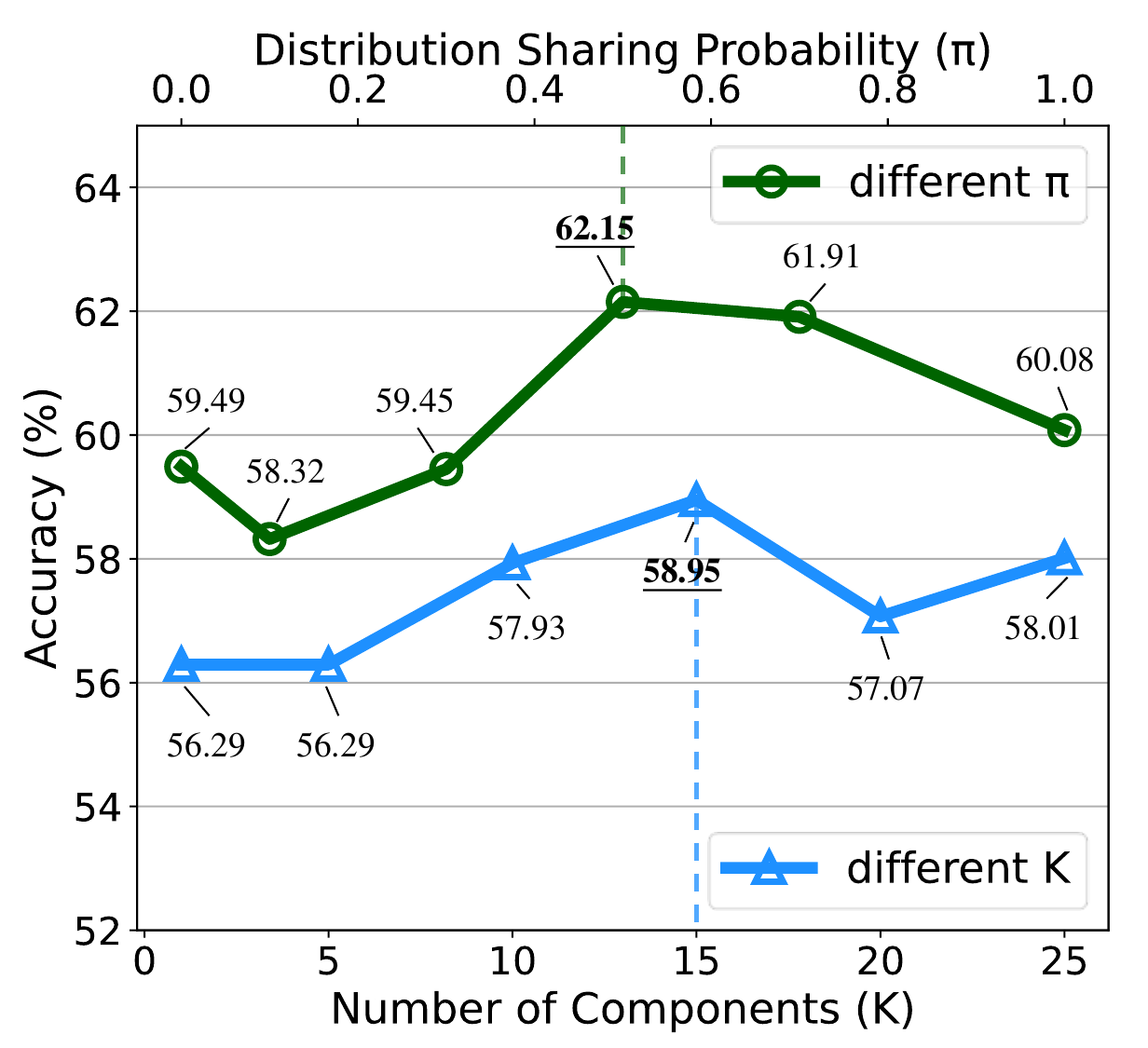}
		\caption{\footnotesize The \textbf{accuracy} (\%) of VIAT-trained ResNet-50 against adversarial viewpoints, using various sets of $K$ and $\pi$.}
		\label{fig:ablation}
	\end{minipage} \hspace{0.2cm}
	\begin{minipage}{0.47\linewidth}
		\centering
		\includegraphics[width=0.99\linewidth]{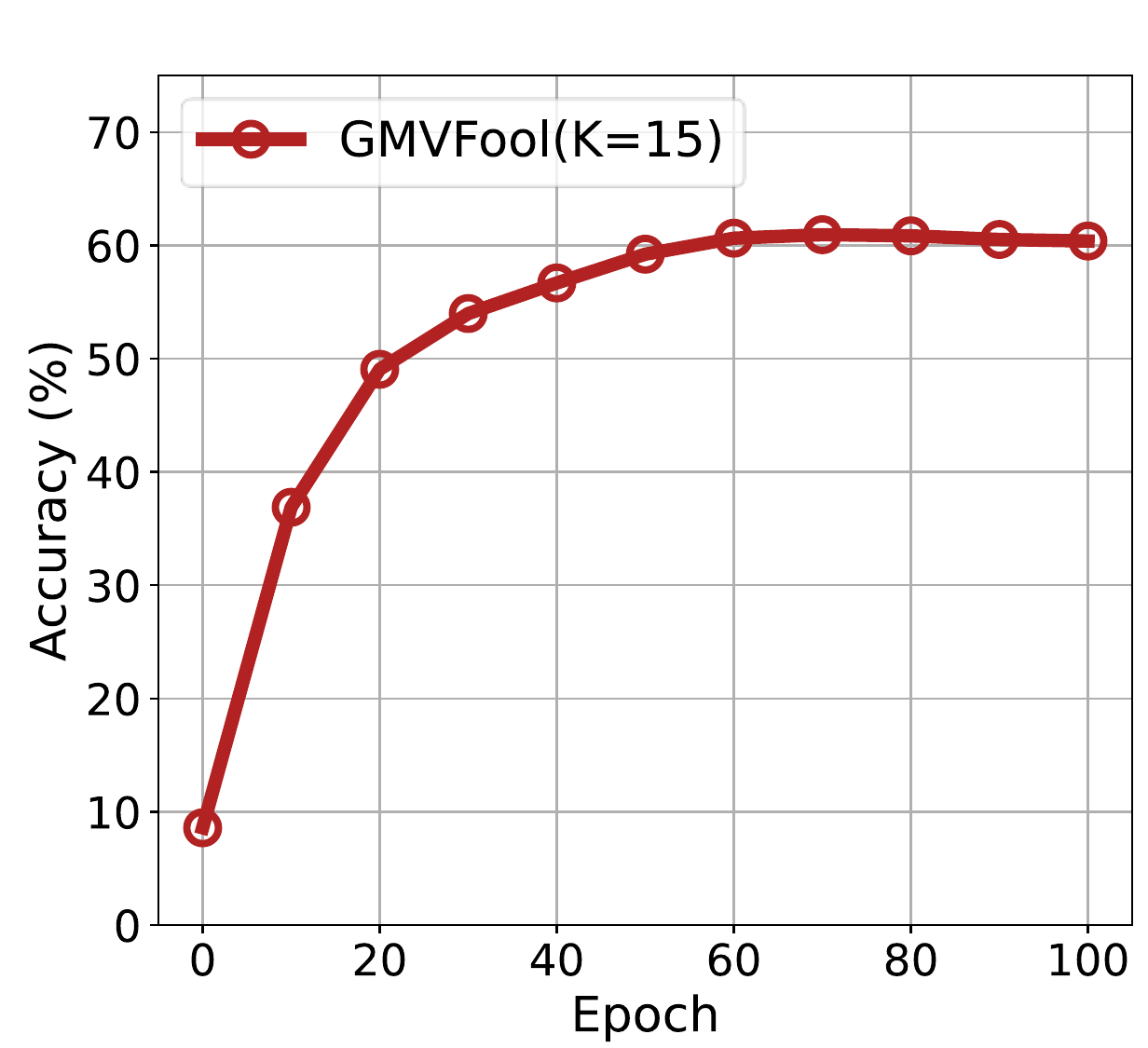}
		\caption{\footnotesize The \textbf{accuracy} (\%) of VIAT-trained ResNet-50 against GMVFool attack with different training iterations.}
		\label{fig:convergence}
	\end{minipage}
\end{figure}

\begin{figure*}[t]
  \centering
  \includegraphics[width=0.96\linewidth]{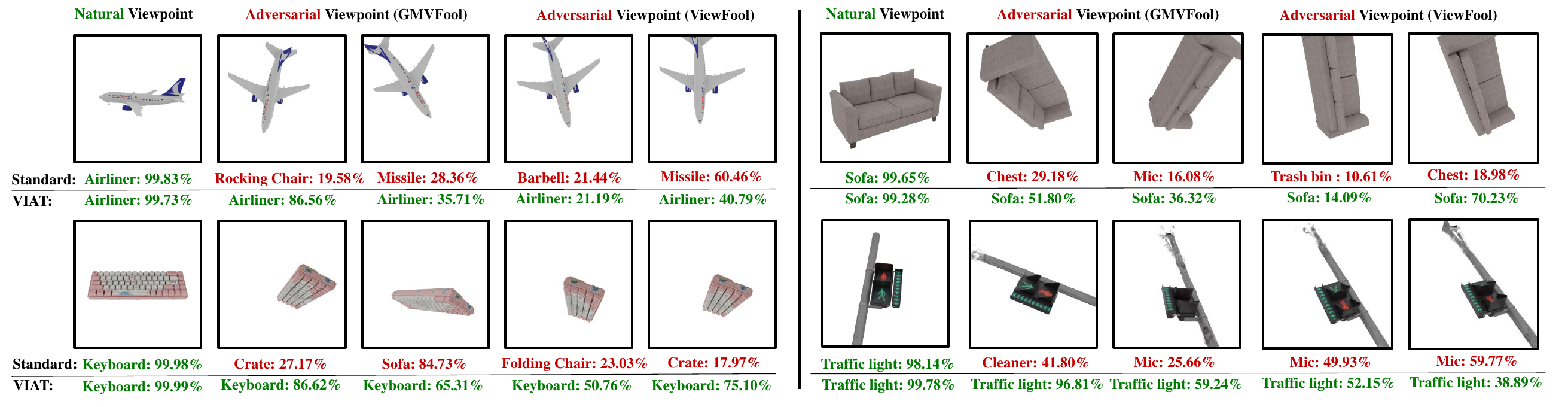}
  \vspace{-0.3cm}
   \caption{The \textbf{prediction} examples of Standard-trained and VIAT-trained ResNet-50 under natural and adversarial viewpoint images. Green and red text represent correct and incorrect predictions, respectively, and the corresponding number is the confidence value.}
   \label{fig:visualization}
\end{figure*}
\begin{figure*}[t]
  \centering

\includegraphics[width=0.95\linewidth]{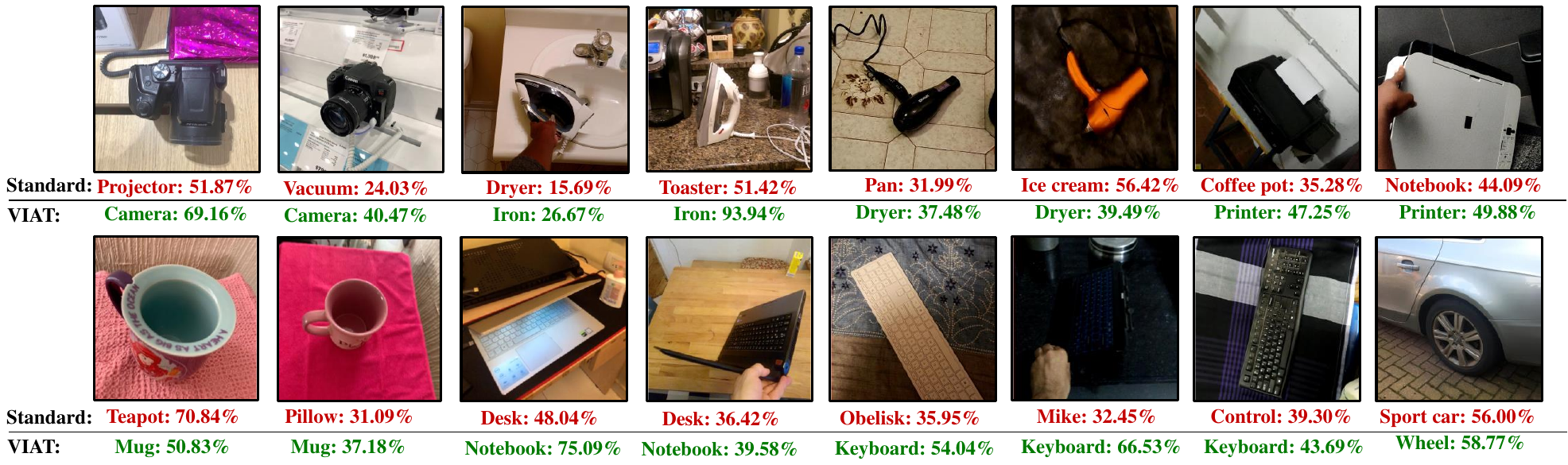}
  \vspace{-0.3cm}
   \caption{The \textbf{prediction} examples of Standard-trained and  VIAT-trained ResNet-50 under the adversarial viewpoint images from real-world objects.}
   \label{fig:phy} 
\end{figure*}

\subsection{Experimental settings} \label{sec:setting}
\noindent \textbf{(A) Training Datasets:} To overcome the scarcity of multi-view images in ImageNet, a new dataset called IM3D is constructed. It comprises 1K typical synthetic 3D objects from 100 ImageNet categories, with each category containing ten objects. We acquire the upper hemisphere's multi-view images and the corresponding camera poses. Subsequently, NeRF representations are learned using Instant-NGP. The objects are further divided into a training set and a validation set, with a ratio of 9:1, facilitating adversarial training and validation of viewpoint invariance. Detailed information about IM3D is available in \textbf{Appendix C}. 

\noindent \textbf{(B) Model:} For our experiments, we consider two classifiers as the model to be attacked, including the CNN-based ResNet-50~\cite{he2016deep} and the Transformer-based ViT-B/16~\cite{dosovitskiy2020image}. We train the classifiers on the subset of ImageNet, which corresponds to our 3D synthetic object’s category until converged. The top-1 accuracies of standard trained ResNet-50 and ViT-B/16 are 85.60\% and 92.88\%, respectively.

\noindent \textbf{(C) VIAT Setting:} Following~\cite{dong2022viewfool}, the virtualized camera is initialize at $[0,4,0]$, then we set the range of rotation parameters  $\psi \in [-180^{\circ}, 180^{\circ}]$, $\theta \in [-30^{\circ}, 30^{\circ}]$, $\phi \in [20^{\circ}, 160^{\circ}]$, the range of translation parameters $\Delta_x \in [-0.5, 0.5]$, $\Delta_y \in [-1, 1]$, $\Delta_z \in [-0.5, 0.5]$, and the balance hyperparameter $\lambda=0.01$. Based on the results of the ablation studies, we choose the number of the components $K$=15 and distribution sharing probability $\pi$=0.5. For the inner maximization, we approximate the gradients in Eq.~\eqref{eq: 12} with $q$=100 MC samples and use the Adam~\cite{kingma2014adam} optimizer to update $\Psi$ for 50 iterations in the first epoch, then iterate ten times under the previous $\Psi$ for subsequent epochs. After obtaining the model trained on the ImageNet subset, we continue to train the model for 60 epochs with the adversarial viewpoints and ImageNet clean samples, with a ratio of 1:32.

\noindent \textbf{(D) Evaluation Metrics/Protocols:} In our experiments, the Top-1 accuracy is employed as the basic evaluation metric. To thoroughly investigate the viewpoint invariance of models, we establish three evaluation protocols: (a) \textbf{ImageNet}: the accuracy is calculated under the validation set of ImageNet. (b) \textbf{ViewFool}: the accuracy is calculated under renderings of adversarial viewpoints generated by ViewFool. (c) \textbf{GMVFool}: the accuracy is calculated under renderings of adversarial viewpoints generated by our GMVFool.


\subsection{Ablation Studies} \label{sec:ablation}

\noindent \textbf{The effects of $K$ and $\pi$.} We conducted ablation experiments to investigate the impact of the number of Gaussian components ($K$) and distribution sharing probability ($\pi$). Fig.~\ref{fig:ablation} illustrates the model's classification accuracy against GMVFool after VIAT training with different settings. We observed a positive correlation within a specific range between the model's resistance to viewpoint attacks and the choice of $K$ in VIAT. It is important to emphasize that the selection of $K$ depends on the specific objectives of GMVFool. Increasing $K$ appropriately as an internal maximization method for VIAT leads to improved robustness. Conversely, smaller $K$ can capture worst-case scenarios for pure viewpoint attacks. The attack performance will be demonstrated in Sec.~\ref{sec:gmvfool performance}. Additionally, a suitable value of $\pi$ contributes to achieving better viewpoint invariance. However, excessively high values of $K$ and $\pi$ can have the opposite effect. In summary, we observed that VIAT performs best when $K\!\!=\!\!15$ and $\pi\!\!=\!\!0.5$.

\noindent \textbf{Convergence with different initial values.} VIAT is a distribution-based adversarial training framework, its convergence is guaranteed in theory~\cite{dong2020adversarial}. Furthermore, We study the convergence of VIAT with a learning rate of 0.001, $K\!\!=\!\!15$, $\pi\!\!=\!\!0.5$. The accuracy under adversarial viewpoints generated by GMVFool is presented in Fig.~\ref{fig:convergence}, indicating that VIAT can converge well under the experimental setting.

\subsection{The Performance of VIAT} \label{sec:4.3}
\textbf{Performance on synthetic objects.} In this experience, We compare the ability of VIAT with three potential baselines: (a) \textbf{Natural}: Data augmentation with the most common viewpoint renderings from training objects’ natural states. For this, we define a range of views frequently appearing in ImageNet for each class (e.g., hotdogs are usually in the top view). (b)  \textbf{Random}: Data augmentation with randomly selected viewpoint rendering of objects in the training set. (c) \textbf{VIAT(ViewFool)}: VIAT but uses ViewFool as the inner maximization method. To be a fair comparison, we also use Instant-NGP to accelerate ViewFool. From the results shown in Table~\ref{table:defense}, we can draw the following conclusions: 

\textbf{\underline{(1)}}
VIAT significantly improves the model's viewpoint invariance. When evaluated under adversarial viewpoints generated by GMVFool and ViewFool, the accuracy of ResNet-50 is improved by 50.63\% and 51.56\%, respectively, while that of ViT-B/16 is improved by 54.03\% and 57.11\%, respectively, compared to the standard-trained model.

\textbf{\underline{(2)}}
Using GMVFool as the inner maximization method for VIAT leads to higher accuracy than using ViewFool under adversarial viewpoint images. The smaller accuracy gap between the two attack methods indicates that VIAT+GMVFool exhibits better generalization to different attacks. This improvement can be attributed to GMVFool's Gaussian mixture modeling, which generates diverse adversarial viewpoints for the network to learn from.

\textbf{\underline{(3)}}
Augmentation methods utilizing natural and random viewpoint images have significant limitations in improving the model's performance under adversarial viewpoints.

\textbf{\underline{(4)}}
ViT-B/16 outperforms ResNet-50 in resisting adversarial viewpoint attacks. This phenomenon may be attributed to its transformer structure, as further confirmed by the benchmark results of ImageNet-V+ in Sec.~\ref{sec:imagenet-v+}.

Fig.~\ref{fig:visualization} shows the visualization of the adversarial viewpoint renderings and the predictions with confidences of the standard-trained and VIAT-trained ResNet-50. The results demonstrate that the VIAT-trained model can maintain the correct prediction when facing an adversarial viewpoint.

\noindent\textbf{Performance on real-world objects.} \label{sec:real-world} We evaluate the proposed methods in the real-world domain. To evaluate the effectiveness of VIAT in handling adversarial viewpoints from real backgrounds, we conduct experiments on Objectron~\cite{ahmadyan2021objectron}, which consists of object-centric videos in the wild. We obtain multi-view images of 5 categories of objects from Objectron, and the prediction results of the Standard-trained and VIAT-trained ResNet-50 models are shown in Table.~\ref{table:objectron}, indicating that the performance of VIAT in enhancing viewpoint robustness can be transferred to real-world objects. Some cases are presented in Fig.~\ref{fig:phy}, where VIAT-trained models maintain correct predictions for significant unnatural viewpoints. 
\begin{table}[t]
\caption{The \textbf{accuracy} of standard-trained and VIAT-trained ResNet-50 in the real-world images from the objectron dataset.}

\label{table:objectron}
\setlength\tabcolsep{5.5pt}
\renewcommand\arraystretch{1.2}
\centering
\begin{tabular}{cl|c|c|c|c|c}
\hline
\multicolumn{2}{c|}{}   & \emph{shoe}           &  \emph{camera}         &  \emph{mug}            &  \emph{chair}          &  \emph{computer}       \\ \hline \hline
\multicolumn{2}{c|}{Standard} & 76.28\%          & 81.42\%          & 97.67\%          & 38.82\%          & 60.19\%          \\
\multicolumn{2}{c|}{VIAT (Ours)}     & \textbf{76.73\%} & \textbf{86.93\%} & \textbf{99.07\%} & \textbf{39.75\%} & \textbf{63.40\%} \\ \hline
\end{tabular}
\end{table}

\begin{table}[t]
\caption{The \textbf{accuracy} of standard-trained, Augmentation with random renderings (Aug) and VIAT-trained ResNet-50 in various OOD and multi-view datasets.} 

\setlength\tabcolsep{3.5pt}
\renewcommand\arraystretch{1.2}
\centering
\begin{tabular}{c|l|ccc}
\hline
                                                                                & Dataset    & Standard &Aug & VIAT (Ours) \\ \hline \hline
\multirow{3}{*}{\begin{tabular}[c]{@{}c@{}}OOD\\ Datasets\end{tabular}}         & ObjectNet~\cite{barbu2019objectnet}  & 35.92\%               & 36.02\%                      & \textbf{37.39\%}  \\  
                                                                                & ImageNet-A~\cite{hendrycks2021natural} & 19.03\%                &  18.65\%                     & \textbf{20.32\%}  \\ 
                                                                                & ImageNet-R~\cite{hendrycks2021many} & 45.06\%               & 45.09\%                      & \textbf{46.51\%}  \\ 
                                                                                & ImageNet-V~\cite{dong2022viewfool} & 28.15\%               & 32.83\%                      & \textbf{38.96\%}  \\ \hline
\multirow{3}{*}{\begin{tabular}[c]{@{}c@{}}Multi-view \\ Datasets\end{tabular}} & MIRO~\cite{kanezaki2018rotationnet}       & 57.41\%                    & 58.78\%                      &  \textbf{65.86\%}               \\ 
                                                                                &
                                                                                OOWL~\cite{ho2019catastrophic}        & 51.41\%                    & 51.24\%                      & \textbf{52.13\%}               \\ 
                                                                                & CO3D~\cite{reizenstein2021common}       & 64.68\%                    &  64.90\%                     &  \textbf{66.04\%}               \\ \hline
\end{tabular}
\vspace{-0.3cm}

\label{table:ood}
\end{table}

\begin{figure}[t]
  \centering

\includegraphics[width=0.95\linewidth]{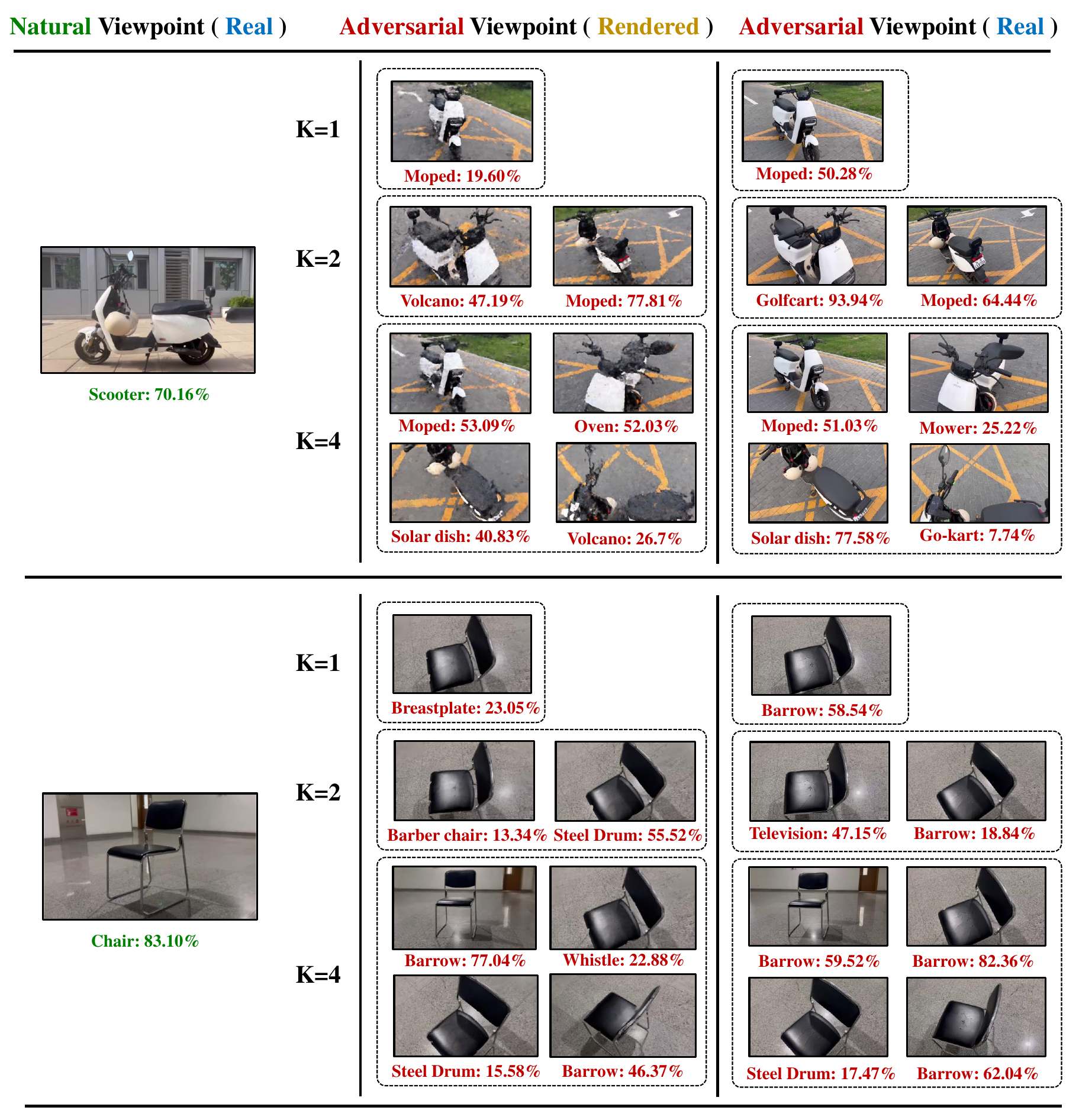}
   \caption{Adversarial viewpoints examples in the Real-World objects, captured by GMVFool with different $K$.}
   \label{fig:phy2} 
\end{figure}

Additionally, following the settings in ViewFool, we record some videos of real-world objects and use GMVFool with different $K$ to capture their adversarial viewpoints. The results in Fig.~\ref{fig:phy2} demonstrate that as $K$ increases, the captured adversarial viewpoints become more diverse, which is directly proportional to the value of K. This observation further confirms the advantage of GMVFool in enhancing the diversity of adversarial viewpoints.

Moreover, we conduct experiments on various \textbf{viewpoint related OOD datasets} and other \textbf{multi-view datasets} that contain viewpoint perturbations. We first select the categories on these datasets that overlap with our training categories and then measure the Top-1 accuracy of our models on these images. The results presented in Table~\ref{table:ood} demonstrate the effectiveness of the VIAT-trained model in resisting natural viewpoint perturbations and showcase its enhanced robustness against viewpoint transformations.

\subsection{The Performance of GMVFool}
\label{sec:gmvfool performance}

\begin{figure}[t]
  \centering
  \includegraphics[width=0.99\linewidth]{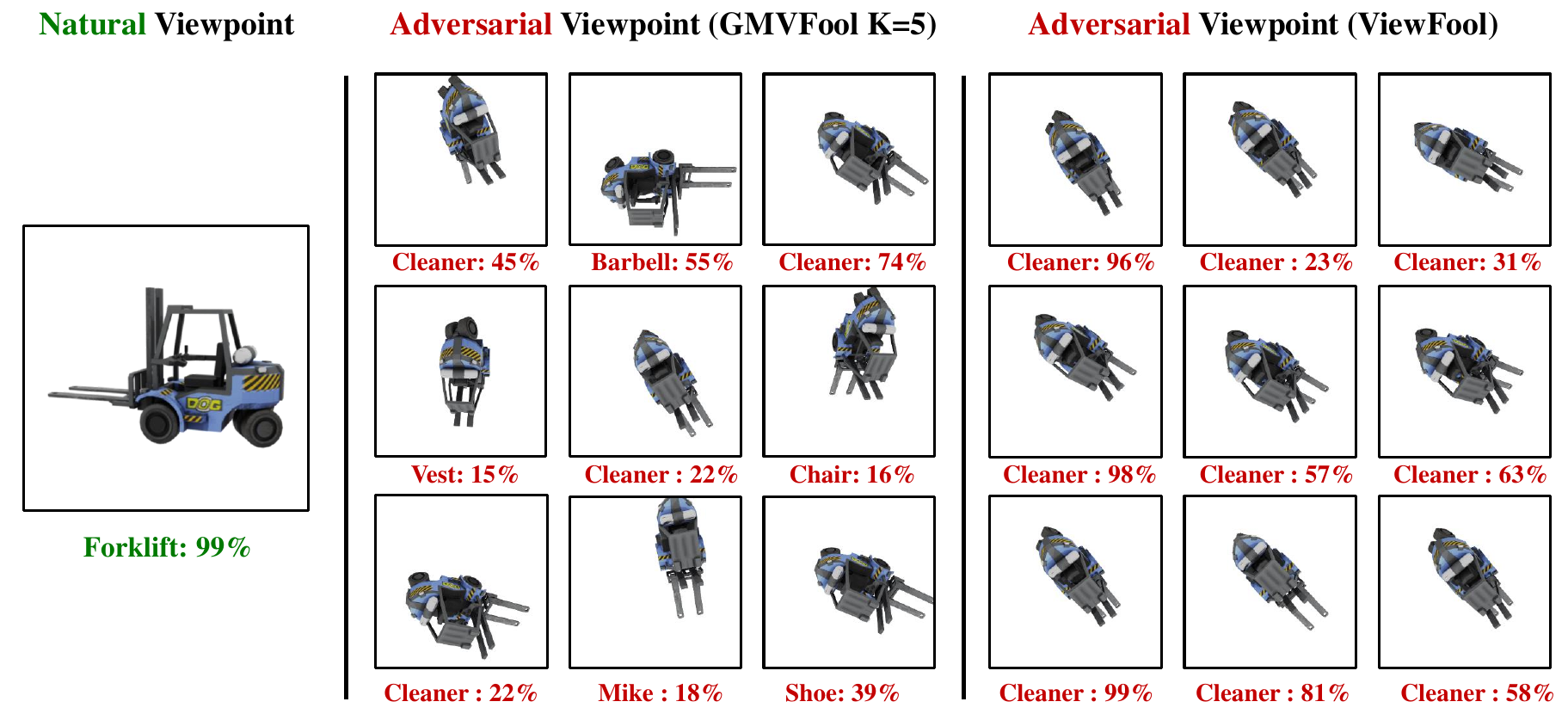} 
   \caption{The comparison of adversarial viewpoints captured by GMVFool  ($K\!\!=\!\!5$)  and ViewFool, and prediction results from ResNet-50.} 
   \label{fig:comparison of adversarial viewpoints}
   \vspace{-0.3cm}
\end{figure}

\begin{figure}[t]
  \centering
  \includegraphics[width=0.95\linewidth]{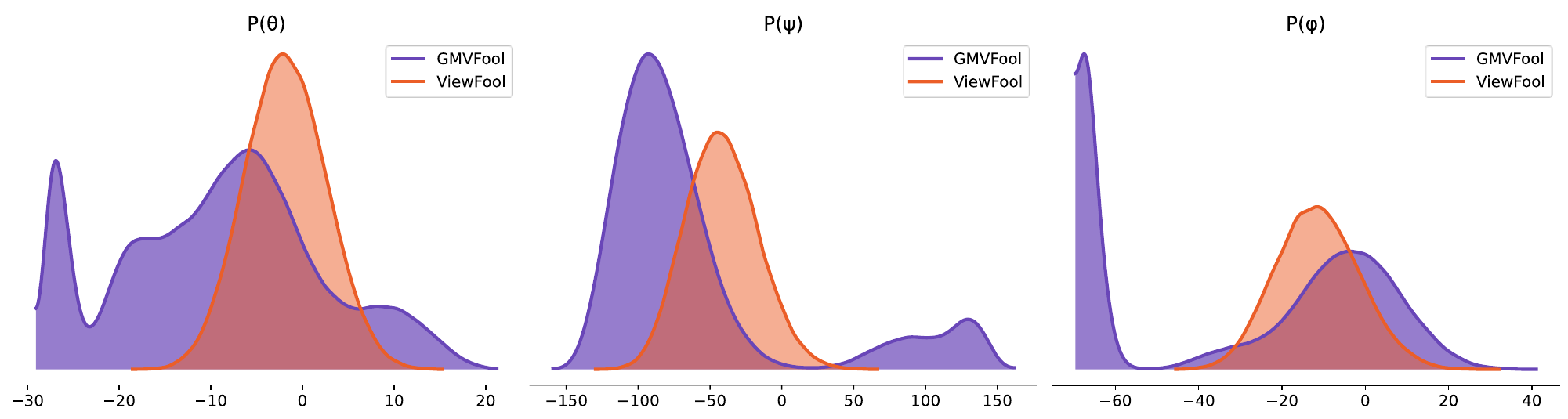} 
   \caption{The \textbf{probability density curves} of the adversarial distribution under viewpoint parameters $\theta$, $\psi$ and $\phi$, which are optimized by GMVFool ($K=5$) and ViewFool, respectively.} 
   \label{fig:distribution}
\end{figure}

\noindent \textbf{Enhanced Diversity.} The mixture distribution design contributes to GMVFool capturing more diverse adversarial viewpoints via a single optimization. We conduct two experiments to illustrate this observation. We plot the Probability Density Curves of the adversarial distribution optimized by GMVFool ($K\!\!=\!\!5$) and ViewFool, as shown in Fig.~\ref{fig:distribution}. It can be observed that compared to the optimal distribution obtained by ViewFool, GMVFool's curve exhibits more peaks, indicating a capture of a more diverse range of adversarial viewpoints. We further sample viewpoints from the adversarial distributions obtained by two methods and render the corresponding images, as shown in Fig.~\ref{fig:comparison of adversarial viewpoints}. These samples successfully deceive ResNet-50. However, the adversarial viewpoint images captured by GMVFool exhibit a higher diversity than ViewFool's. This characteristic empowers us to understand the "blind spots" of models comprehensively and benefits adversarial training.

\begin{figure}[t]
  \centering

\includegraphics[width=0.95\linewidth]{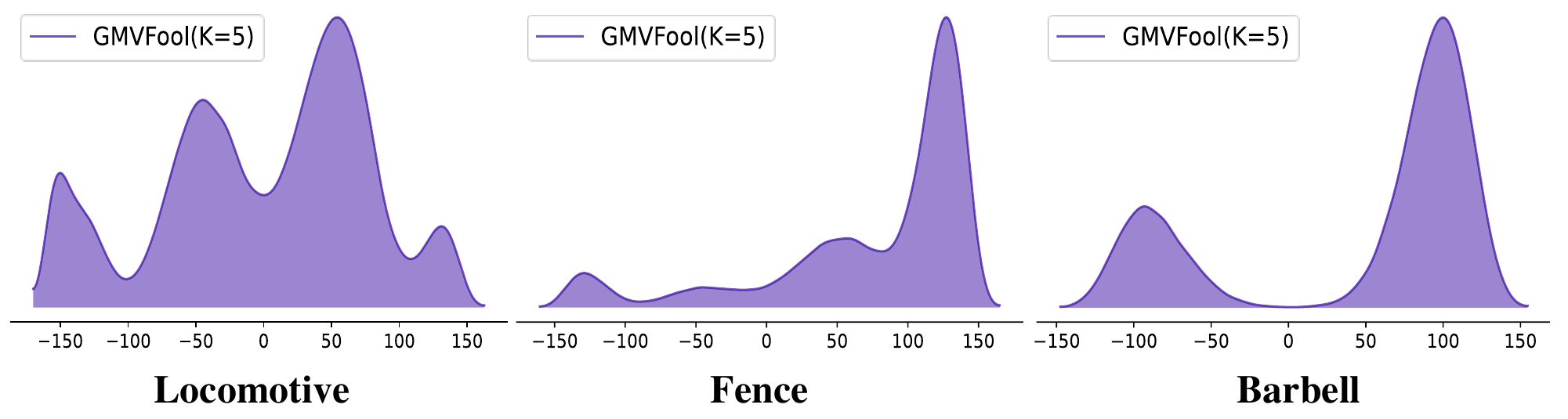}
   \caption{The optimal distribution across categories by GMVFool ($K\!\!=\!\!5$)}
   \label{fig:dist_2} 
\vspace{-0.3cm}
\end{figure}



\noindent \textbf{Adaptive Degradation.} Since GMVFool optimizing the component weights $\omega_k$, the mixture distribution possesses adaptive degradation capability, i.e., for objects with limited adversarial viewpoints, redundant components will overlap or degrade during optimization. Therefore, the choice of $K$ actually determines the upper bound of adversarial viewpoint diversity. As shown in Fig.~\ref{fig:dist_2}, the optimal distribution across categories is adaptive despite setting the same $K$ for GMVFool. This capability considers the category-dependence of the adversarial viewpoints number and enables GMVFool to adjust the mixture distribution based on the specific object flexibly.

\begin{table*}[t]
\caption{The \textbf{attack success rate} and the \textbf{entropy} of methods against various classifiers. $\mathcal{R}(\cdot)$ denotes the rendering process.}
\setlength\tabcolsep{3.8pt}
\renewcommand\arraystretch{1.2}
\centering
\begin{tabular}{c|cc|cc|cc|cc}
\hline
\multirow{2}{*}{Method} & \multicolumn{2}{c|}{ResNet-50}                                                       & \multicolumn{2}{c|}{EN-B0}                                                           & \multicolumn{2}{c|}{DeiT-B}                                                          & \multicolumn{2}{c}{Swin-B}                                                           \\ \cline{2-9} 
                        & \multicolumn{1}{c|}{$\mathcal{R}(p^*(\mathbf{v}))\uparrow$} & $\mathcal{H}(p^*(\mathbf{v}))\uparrow$ & \multicolumn{1}{c|}{$\mathcal{R}(p^*(\mathbf{v}))\uparrow$} & $\mathcal{H}(p^*(\mathbf{v}))\uparrow$ & \multicolumn{1}{c|}{$\mathcal{R}(p^*(\mathbf{v}))\uparrow$} & $\mathcal{H}(p^*(\mathbf{v}))\uparrow$ & \multicolumn{1}{c|}{$\mathcal{R}(p^*(\mathbf{v}))\uparrow$} & $\mathcal{H}(p^*(\mathbf{v}))\uparrow$ \\ \hline\hline
Random Search           & \multicolumn{1}{c|}{64.12\%}                          & -                              & \multicolumn{1}{c|}{79.18\%}                          & -                              & \multicolumn{1}{c|}{32.25\%}                          & -                              & \multicolumn{1}{c|}{19.88\%}                          & -                              \\
ViewFool                & \multicolumn{1}{c|}{91.50\%}                          & -10.14                         & \multicolumn{1}{c|}{95.64\%}                          & -10.38                         & \multicolumn{1}{c|}{80.37\%}                          & -10.41                         & \multicolumn{1}{c|}{73.85\%}                          & -10.71                         \\ \hline
GMVFool (K=1)            & \multicolumn{1}{c|}{\textbf{92.13\%}}                 & -10.17                         & \multicolumn{1}{c|}{\textbf{95.78\%}}                 & -10.38                         & \multicolumn{1}{c|}{\textbf{80.61\%}}                 & -10.44                         & \multicolumn{1}{c|}{\textbf{74.02\%}}                 & -10.62                         \\
GMVFool (K=3)            & \multicolumn{1}{c|}{89.20\%}                          & -3.75                          & \multicolumn{1}{c|}{\textbf{95.78\%}}                          & -3.94                          & \multicolumn{1}{c|}{79.65\%}                          & -3.92                          & \multicolumn{1}{c|}{72.48\%}                          & -4.05                          \\
GMVFool (K=5)            & \multicolumn{1}{c|}{91.56\%}                          & \textbf{-0.69}                 & \multicolumn{1}{c|}{95.62\%}                          & \textbf{-0.85}                 & \multicolumn{1}{c|}{78.91\%}                          & \textbf{-1.00}                 & \multicolumn{1}{c|}{70.16\%}                          & \textbf{-1.07}                 \\ \hline
\end{tabular}

\label{table:attack}
\end{table*}

\begin{table*}[t]
\caption{The \textbf{accuracy} of standard-trained and VIAT-trained models under ImageNet-V+ dataset.}
\setlength\tabcolsep{5.5pt}
\renewcommand\arraystretch{1.2}
\centering
\begin{tabular}{c|ccccc|ccc}
\hline
\multirow{2}{*}{}             & \multicolumn{5}{c|}{CNN-based}                                                                                                                                     & \multicolumn{3}{c}{Transformer-based}                                  \\ \cline{2-9} 
                              & \multicolumn{1}{c|}{ResNet-50} & \multicolumn{1}{c|}{Inception-v3} & \multicolumn{1}{c|}{Inc-Res-v2} & \multicolumn{1}{c|}{EffcientNet-B0} & DenseNet-121 & \multicolumn{1}{c|}{ViT-B/16} & \multicolumn{1}{c|}{DeiT-B}  & Swin-B  \\ \hline \hline
Standard-trained              & \multicolumn{1}{c|}{15.74\%}   & \multicolumn{1}{c|}{11.33\%}      & \multicolumn{1}{c|}{12.61\%}             & \multicolumn{1}{c|}{14.27\%}        & 13.90\%      & \multicolumn{1}{c|}{11.80\%}  & \multicolumn{1}{c|}{17.46\%} & 24.00\% \\ \hline
\multirow{2}{*}{VIAT-trained} & \multicolumn{1}{c|}{31.28\%}   & \multicolumn{1}{c|}{35.68\%}      & \multicolumn{1}{c|}{26.89\%}             & \multicolumn{1}{c|}{33.79\%}        & 34.53\%      & \multicolumn{1}{c|}{75.49\%}  & \multicolumn{1}{c|}{61.09\%} & 79.25\% \\
                              & \multicolumn{1}{c|}{($\uparrow$~15.54\%)}   & \multicolumn{1}{c|}{($\uparrow$~24.35\%)}      & \multicolumn{1}{c|}{($\uparrow$~14.28\%)}             & \multicolumn{1}{c|}{($\uparrow$~19.52\%)}        & $(\uparrow$~20.63\%)      & \multicolumn{1}{c|}{($\uparrow$~63.69\%)}  & \multicolumn{1}{c|}{($\uparrow$~43.63\%)} & ($\uparrow$~55.25\%) \\ \hline
\end{tabular}
\label{table:benchmark}
\end{table*}

\begin{table}[htb]
\caption{The \textbf{accuracy} of VLMs under ImageNet, IM3D and ImageNet-V+ dataset.}
\setlength\tabcolsep{3.0pt}
\renewcommand\arraystretch{1.2}
\centering
\begin{tabular}{c|c|c|c|c|c}
\hline
Models                  & \begin{tabular}[c]{@{}c@{}}Vision\\ Encoder\end{tabular} & \begin{tabular}[c]{@{}c@{}}\#~Toatal\\ Params\end{tabular} & \scriptsize{ImageNet}                & IM3D     & \scriptsize{ImageNet-V+} \\ \hline \hline
\multirow{2}{*}{\scriptsize ALFBEF~\cite{li2021align}} & \multirow{2}{*}{ViT-B/16}                               & \multirow{2}{*}{210M}                                   & \multirow{2}{*}{66.06\%} & 52.88\%   & 26.22\%      \\
                        &                                                         &                                                           &                         & ($\downarrow$~13.18\%) & ($\downarrow$~39.84\%)    \\ \hline
\multirow{8}{*}{\scriptsize CLIP~\cite{radford2021learning}}   & \multirow{2}{*}{ResNet-50}                              & \multirow{2}{*}{102M}                                   & \multirow{2}{*}{65.12\%} & 53.36\%   & 2.51\%       \\ 
                        &                                                         &                                                           &                         & ($\downarrow$~11.76\%) & ($\downarrow$~62.61\%)    \\ \cline{2-6} 
                        & \multirow{2}{*}{ViT-B/16}                               & \multirow{2}{*}{150M}                                   & \multirow{2}{*}{76.94\%} & 66.60\%   & 37.99\%      \\ 
                        &                                                         &                                                           &                         & ($\downarrow$~10.34\%) & ($\downarrow$~38.95\%)    \\ \cline{2-6} 
                        & \multirow{2}{*}{ViT-B/32}                               & \multirow{2}{*}{151M}                                   & \multirow{2}{*}{72.74\%} & 58.59\%   & 29.24\%      \\
                        &                                                         &                                                           &                         & ($\downarrow$~14.15\%) & ($\downarrow$~43.50\%)    \\ \cline{2-6} 
                        & \multirow{2}{*}{ViT-L/14}                               & \multirow{2}{*}{428M}                                   & \multirow{2}{*}{81.96\%} & 76.16\%   & 48.49\%      \\
                        &                                                         &                                                           &                         & ($\downarrow$~5.80\%)  & ($\downarrow$~33.47\%)    \\ \hline
\multirow{2}{*}{\scriptsize BLIP~\cite{li2022blip}}   & \multirow{2}{*}{ViT-B/16}                               & \multirow{2}{*}{224M}                                   & \multirow{2}{*}{70.02\%} & 70.73\%   & 40.08\%      \\
                        &                                                         &                                                           &                         & ($\uparrow$~0.71\%)  & ($\downarrow$~29.94\%)    \\ \hline
\multirow{4}{*}{\scriptsize BLIP-2~\cite{li2023blip}} & \multirow{2}{*}{ViT-L/14}                               & \multirow{2}{*}{449M}                                   & \multirow{2}{*}{73.86\%} & 76.38\%   & 50.05\%      \\
                        &                                                         &                                                           &                         & ($\uparrow$~2.52\%)  & ($\downarrow$~23.81\%)    \\ \cline{2-6} 
                        & \multirow{2}{*}{ViT-G/14}                               & \multirow{2}{*}{1.2B}                                    & \multirow{2}{*}{77.40\%} & 83.76\%   & 57.92\%      \\
                        &                                                         &                                                           &                         & ($\uparrow$~6.36\%)  & ($\downarrow$~19.48\%)    \\ \hline
\end{tabular}
\label{table:vlp}
\end{table}

\noindent \textbf{Improved Efficiency.} GMVFool achieves efficient optimization of adversarial viewpoints. We compare the time consumption of ViewFool and GMVFool in optimizing the adversarial viewpoint distribution for a single object. Utilizing a single NVIDIA V100 GPU, ViewFool requires \textbf{8 hours} for the optimization process, whereas GMVFool completes the task in only \textbf{3 minutes}, making it nearly \textbf{160×} faster than ViewFool. This improvement in efficiency holds great practical significance as it enables the rapid generation of diverse and aggressive adversarial viewpoints, thereby facilitating efficient adversarial training.

\noindent \textbf{Attack Performance.} GMVFool enhances the diversity of adversarial viewpoints while maintaining a strong attack performance. We evaluate the sampling attack success rate and distribution entropy of the viewpoint attack method against different classifiers in Table~\ref{table:attack}. To ensure a fair comparison of attack performance, as discussed in Sec.~\ref{sec:ablation}, we use smaller values for $K$. The results indicate that GMVFool ($K\!\!=\!\!1$) achieves the highest attack success rate. Remarkably, when $K\!\!=\!\!3$ and $5$, the distribution entropy significantly increased while maintaining a high attack success rate. This signifies that GMVFool successfully captured a broader range of adversarial viewpoints.

\subsection{ImageNet-V+ Benchmark}
\label{sec:imagenet-v+}

To comprehensively evaluate the viewpoint robustness of visual recognition models, we construct a large-scale benchmark dataset: ImageNet-V+. It consists of 100k adversarial viewpoint images generated using GMVFool against ViT-B/16. These adversarial viewpoints are derived from 1k synthetic 3D objects from IM3D that overlap with 100 ImageNet classes. The details and visualizations of ImageNet-V+ will be included in the \textbf{Appendix B}. Then, we report the evaluation results of various models under ImageNet-V+.

\noindent \textbf{Benchmarking traditional architecture models.} We adopt ImageNet-V+ to evaluate 40 different models pre-trained on ImageNet, including models with different structures (VGG~\cite{simonyan2014very}, ResNet~\cite{he2016deep}, Inception~\cite{szegedy2017inception}, DenseNet~\cite{huang2017densely}, EfficientNet~\cite{tan2019efficientnet}, MobileNet-v2~\cite{sandler2018mobilenetv2}, ViT~\cite{dosovitskiy2020image}, DeiT~\cite{touvron2021training}, Swin Transformer~\cite{liu2021swin}, and the MLP Mixer~\cite{tolstikhin2021mlp}), different training paradigms (adversarial training~\cite{salman2020adversarially}, mask-autoencoder~\cite{he2022masked}), different augmentation methods (AugMix~\cite{hendrycks2019augmix}, DeepAugment~\cite{hendrycks2021many}). For comparison, we also compare the model trained with VIAT.  

\begin{figure*}[t]
  \centering
  \includegraphics[width=0.99\linewidth]{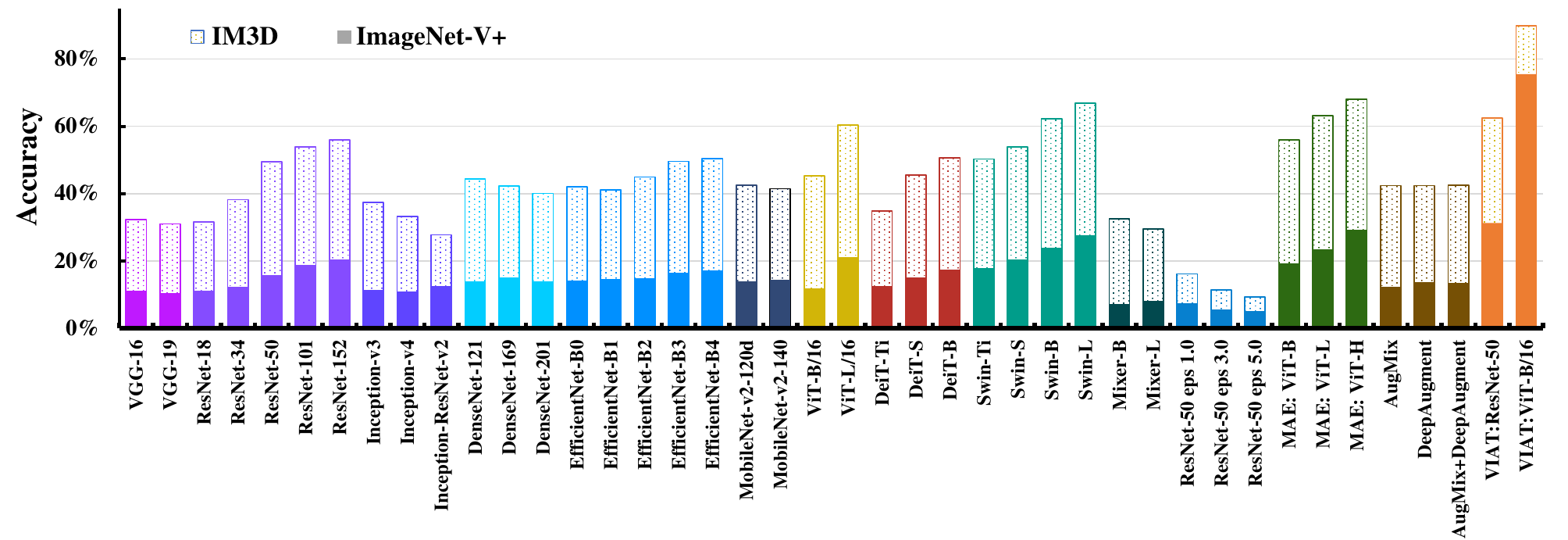}
\caption{The \textbf{accuracy} of different classifiers on natural viewpoint and on ImageNet-V+.}
   \label{fig:benchmark}
\end{figure*} 

\begin{table*}[t]
\caption{The \textbf{average certification radius (ACR)} and \textbf{certification accuracy (CA)} of standard-trained and VIAT-trained models under IM3D.}
\setlength\tabcolsep{5.3pt}
\renewcommand\arraystretch{1.2}
\centering
\begin{tabular}{c|c|ccccc|ccc}
\hline
\multirow{2}{*}{Metric} & \multirow{2}{*}{Method}       & \multicolumn{5}{c|}{CNN-based}                                                                                                                                     & \multicolumn{3}{c}{Transformer-based}                                \\ \cline{3-10} 
                        &                               & \multicolumn{1}{c|}{ResNet-50} & \multicolumn{1}{c|}{Inception-v3} & \multicolumn{1}{c|}{Inc-Res-v2} & \multicolumn{1}{c|}{EffcientNet-B0} & DenseNet-121 & \multicolumn{1}{c|}{ViT-B/16} & \multicolumn{1}{c|}{DeiT-B} & Swin-B \\ \hline\hline
\multirow{3}{*}{ACR~$(\uparrow)$}    & Standard-trained              & \multicolumn{1}{c|}{0.105}          & \multicolumn{1}{c|}{0.085}             & \multicolumn{1}{c|}{0.079}                    & \multicolumn{1}{c|}{0.081}               & 0.097             & \multicolumn{1}{c|}{0.167}         & \multicolumn{1}{c|}{0.158}       & 0.184        \\ \cline{2-10} 
                        & \multirow{2}{*}{VIAT-trained} & \multicolumn{1}{c|}{0.150}          & \multicolumn{1}{c|}{0.131}             & \multicolumn{1}{c|}{0.112}                    & \multicolumn{1}{c|}{0.129}               & 0.132             & \multicolumn{1}{c|}{0.201}         & \multicolumn{1}{c|}{0.183}       & 0.193        \\
                        &                            & \multicolumn{1}{c|}{($\uparrow$~0.045) }          & \multicolumn{1}{c|}{($\uparrow$~0.046)}             & \multicolumn{1}{c|}{($\uparrow$~0.033)}                    & \multicolumn{1}{c|}{($\uparrow$~0.048)}               &  ($\uparrow$~0.035)            & \multicolumn{1}{c|}{($\uparrow$~0.034)}         & \multicolumn{1}{c|}{($\uparrow$~0.025)}       & ($\uparrow$~0.009)        \\ \hline
\multirow{3}{*}{CA~$(\uparrow)$}     & Standard-trained              & \multicolumn{1}{c|}{41\%}          & \multicolumn{1}{c|}{29\%}             & \multicolumn{1}{c|}{32\%}                    & \multicolumn{1}{c|}{24\%}               & 31\%            & \multicolumn{1}{c|}{78\%}         & \multicolumn{1}{c|}{78\%}       &85\%        \\ \cline{2-10} 
                        & \multirow{2}{*}{VIAT-trained} & \multicolumn{1}{c|}{77\%}          & \multicolumn{1}{c|}{61\%}             & \multicolumn{1}{c|}{61\%}                    & \multicolumn{1}{c|}{66\%}               &  67\%            & \multicolumn{1}{c|}{94\%}         & \multicolumn{1}{c|}{91\%}       &  90\%      \\
                        &                               & \multicolumn{1}{c|}{($\uparrow$~36\%)  }          & \multicolumn{1}{c|}{($\uparrow$~32\%) }             & \multicolumn{1}{c|}{($\uparrow$~29\%) }                    & \multicolumn{1}{c|}{($\uparrow$~42\%) }               & ($\uparrow$~36\%)               & \multicolumn{1}{c|}{($\uparrow$~16\%) }         & \multicolumn{1}{c|}{($\uparrow$~13\%) }       & ($\uparrow$~5\%)        \\ \hline
\end{tabular}
\label{table:certify}
\end{table*}

Fig.~\ref{fig:benchmark} illustrates the accuracy of various models on natural viewpoint images and ImageNet-V+. When exposed to adversarial viewpoints, the accuracy of all models decreases significantly. We observe that the model's performance with the same architectures is positively related to its size, with transformer-based models outperforming CNN-based models. Among them, MAE with ViT-H performs best in ImageNet-V+, achieving 29.37\% accuracy. Models using data augmentation and adversarial training, which is robust to adversarial examples and image corruption in previous work, perform poorly from the adversarial viewpoint. We further select eight models with different architectures and train them using VIAT. The changes in classification accuracy on ImageNet-V+ are reported in Table~\ref{table:benchmark}; all models exhibit varying degrees of improvement in performance on ImageNet-V+. Among them, Transformer-based models show a significant increase compared to the CNN-based models. Particularly, the VIAT-trained Swin-B outperformed other models, achieving an accuracy of 79.25\%.

\noindent \textbf{Benchmarking the large vision-language models.} Recent studies have showcased the remarkable performance of Large Vision-Language Models (VLMs) across various visual tasks. Using pre-trained VLMs as zero-shot classifiers surpasses traditional models' accuracy on standard datasets and demonstrate impressive OOD robustness on benchmarks like ImageNet-A and ImageNet-R~\cite{radford2021learning}. However, the viewpoint robustness of VLMs has yet to receive a comprehensive evaluation. Thus, we present the zero-shot classification results of representative VLMs on ImageNet-V+. Table~\ref{table:vlp} reveals that although VLMs outperform traditional models significantly under natural viewpoints (ImageNet) and multi-view scenarios (IM3D), they exhibit noticeable performance degradation when confronted with adversarial viewpoints. Notably, CLIP (ResNet-50) has the most substantial accuracy drop (62.61\%), while BLIP (VIT-G) demonstrates the least degradation (19.48\%). Moreover, we observe that models with larger parameter sizes exhibit better viewpoint robustness. We also observe the significant impact of adversarial viewpoints on VLMs' performance in the VQA task, as shown in \textbf{Appendix B}. Considering these findings, we emphasize the need to enhance the viewpoint robustness of VLMs as a crucial task for future research.

\subsection{Certification Results of ViewRS}
\noindent \textbf{Setup.} In experience, we estimate the certification radius for each object using 1,000 Monte Carlo samples with a failure probability of $10^{-3}$. We set the initial viewpoint of all objects to $\mathbf{v}\!\! = \!\![ 0,0,65^{\circ},0, 0, 0 ]$, which corresponds to the natural viewpoint of most objects. We sample viewpoint perturbations using Gaussian noise with a variance $\tilde{\sigma}\!\!=\!\!0.1$. Finally, we obtained various models' Average Certification Radius (ACR) and Certification Accuracy (CA) under the IM3D test set. To provide a sense of scale, a viewpoint parameter perturbation with certification radius 0.1 could change initial viewpoint by $[\pm3^{\circ},\pm18^{\circ}, \pm7^{\circ},\pm 0.05,\pm 0.1,\pm 0.05]$. Based on the results in Table~\ref{table:certify}, we draw the following conclusions:

\noindent \textbf{Transformers perform remarkably well}. Regarding ACR and CA, we observe that transformer-based models exhibit stronger certified robustness against viewpoint changes than CNN-based models, consistent with the conclusions in Sec.~\ref{sec:imagenet-v+}. Specifically, Swin-B achieves the best ACR and CA among the standard-trained models. However, among the models trained with VIAT, ViT-B performs the best.

\noindent \textbf{VIAT is effective}. After training with VIAT, the models show varying degrees of improvement in terms of ACR and CA, which means the constructed smooth classifiers are capable of withstanding stronger viewpoint perturbations. This result is further support the effectiveness of VIAT in enhancing the viewpoint robustness.  Unlike the empirical evaluations mentioned above, the robustness observed here is certifiable. Furthermore, we find that VIAT achieves significantly higher improvements for CNN models compared to Transformer models.
\label{sec:Certification}

\section{Conclusion}
\label{sec:5}
In this paper, we proposed a novel and effective framework: VIAT, which leveraged adversarial training to achieve viewpoint-invariant visual models. We also designed GMVFool, an efficient method for generating diverse adversarial viewpoints. To validate the effectiveness of VIAT in enhancing viewpoint invariance, we contributed ImageNet-V+, a large-scale viewpoint OOD benchmark containing 100k adversarial viewpoint images generated by GMVFool. We provided evaluation results of traditional models and large vision-language models on ImageNet-V+. To further demonstrate the effectiveness of VIAT, we presented a viewpoint robustness certification method specifically for visual models and compared the certification results between VIAT and standard training.





\ifCLASSOPTIONcaptionsoff
  \newpage
\fi



%

\bibliographystyle{splncs04}
\bibliography{egbib}

\end{document}